%% file: rubicon.tex
\documentclass[10pt,a4paper]{article}

\usepackage[T1]{fontenc}
\usepackage{graphicx}

\usepackage{listings}

\usepackage{color}
\usepackage[utf8]{inputenc}
\usepackage{authblk}

\definecolor{dkgreen}{rgb}{0,0.6,0}
\definecolor{gray}{rgb}{0.5,0.5,0.5}
\definecolor{mauve}{rgb}{0.58,0,0.82}

\newcommand\BibTeX{{\rmfamily B\kern-.05em \textsc{i\kern-.025em b}\kern-.08em
T\kern-.1667em\lower.7ex\hbox{E}\kern-.125emX}}
\providecommand{\keywords}[1]{\textbf{\textit{Keywords - }} #1}

\newcommand{\comm}{Communication Layer }
\newcommand{\learn}{Learning Layer }
\newcommand{\prog}{RUBICON }

\begin{document}

\title{A Communication Layer for Integrated Sensors and Robotic ecology Solutions to Ambient Intelligence}

\author[1]{Giuseppe Amato}
\author[2]{Stefano Chessa}
\author[3]{Mauro Dragone}
\author[1]{Claudio Gennaro\thanks{Corresponding author: Claudio Gennaro, claudio.gennaro@isti.cnr.it}}
\author[1]{Claudio Vairo}
\affil[1]{ISTI-CNR, Via G. Moruzzi 1, 56124 Pisa, Italy}
\affil[2]{Department of Computer Science, University Pisa, Largo B. Pontecorvo 1, 56127, Pisa, Italy}
\affil[3]{School of Engineering \& Physical Sciences; Sensors, Signals \& Systems, Heriot-Watt University, Edinburgh, EH14 4AS, UK}

\date{}

\maketitle


\begin{abstract}
This paper presents a communication framework built to simplify the construction of robotic ecologies, i.e., networks of heterogeneous computational nodes interfaced with sensors, actuators, and mobile robots. Building integrated ambient intelligence (AmI) solutions out of such a wide range of heterogeneous devices is a key requirement for a range of application domains, such as home automation, logistic, security and Ambient Assisted Living (AAL). This goal is challenging since these ecologies need to adapt to changing environments and especially when they include tiny embedded devices with limited computational resources. We discuss a number of requirements characterizing this type of systems and illustrate how they have been addressed in the design of the new communication framework. The most distinguishing aspect of our frameworks is the transparency with which the same communication features are offered across heterogeneous programming languages and operating systems under a consistent API. Finally, we illustrate how the framework has been used to bind together and to support the operations of all the components of adaptive robotic ecologies in two real-world test-beds.

\end{abstract}


\section{Introduction}
Wireless Sensor Networks (WSNs) \cite{baronti2007wireless} are an important enabling technology for Ambient Intelligence (AmI), since their capillary capability for sensing the environment provides essential context information to AmI applications.

In conventional approaches, a WSN comprises a number of wireless devices (called sensors, for simplicity), each of which is a micro-system embedding an heterogeneous set of transducers and a radio transceiver. The sensors cooperate to form the (wireless) network, to sense the environment and to collect sensed data in a common gateway that, in turn, makes sensed data available to external applications and users. However, in most recent approaches, the scope of WSN has gone beyond the basic data sensing, collection and aggregation tasks, to include also complex data fusion and analysis. Following this trend, recent projects \cite{bacciu2013distributed,moustapha2008wireless,li2008detecting,nakano2008synchronization} introduced neural computation capabilities in WSNs. In this manner, the sensors can process in real time the data streams of sensed data to recognize complex events occurring in the environment, despite noisy and/or imprecise sensor information.
This feature is of extreme importance to AmI applications in general, since WSN enriched with neural computation functionalities may learn to recognize events that are specific of the environment where they are deployed, and they thus may provide such capability to any mobile device visiting or operating in the same environment. 

In particular, Robotics research has recently demonstrated the advantages of incorporating robots as part of AmI solutions.
The resulting systems can provide complex services with their ability to acquire and apply knowledge about their environment by fusing information from a variety of sources, use intelligence to determine a best course of action, interact with humans, and exploit the flexibility given by robots’ mobility and their robotic skills to interact with the physical world.
A key challenge in this type of solutions is how to make them self-adaptive, so that they can be more easily applied to real-world settings without requiring costly human supervision and configuration every time they need to be tailored to
different environments and configured to suit the needs of their users.

This is the focus of the EU FP7 project RUBICON, (Robotic UBIquitous COgnitive Network) \cite{amato2012robotic,dragone2015cognitive,amato2015robotic}, which has built robotic ecologies consisting of mobile robotic devices, sensors, effectors, and appliances cooperating to perform complex tasks such as supporting elders to live independently. While the emphasis in RUBICON is providing robotic ecologies with self-adaptation features by equipping them with cognitive abilities, such as learning and planning, this paper focuses on the underlying communication capabilities required by this type of  solutions. These capabilities should support flexible communications among components deployed on heterogeneous and robotic devices, and also facilitate the information flow among distributed learning components.


Unfortunately, existing robot/WSN communication approaches exploit ad-hoc communication mechanisms and do not support learning processes, which limit the range of applications to which they can be applied. In addition, at present, it is still too difficult to build actual applications, as this requires many researchers to program each part of the system, using different abstractions, languages and tools, before systems prototypes can be evaluated, usually in laboratory settings.

In contrast to previous solutions, our communication framework is built on top of state of the art robotic and WSN middleware and it is purposefully designed to address the characteristic requirements dictated by adaptive robotic ecologies.
Noticeably, our communication framework is general and allows one to build different applications, all involving the integration of the actuation capabilities of autonomous mobile robots and the sensing and learning capabilities of the devices. 

In earlier papers \cite{vairo2010modeling,amato2011efficient,amato2012when}, we presented a preliminary version of our communication framework. The version presented in this paper provides a more generic architecture of messaging that (i) abstracts from the underlying communication means/protocols, (ii) includes mechanisms to support message reliability and concurrency safety, and (iii) facilitates the management of open systems, in which new devices can join while others can leave the network, at run-time.
Crucially, the framework illustrated in this paper offers the same features and the same application programming interface (API) across both tiny embedded WSN nodes and more capable devices. This hides the complexity that stems from the heterogeneity of the devices and programming languages, and simplifies the development of integrated, interoperable and extensible systems.

The remainder of the paper is organized in the following
manner: Section \ref{sec:relwork} provides an overview of the
most significant robot/WSN integration approaches attempted in
past research, focusing on how they have addressed the
communication requirements. Section \ref{sec:requirements}
provides a general overview of the proposed communication framework and discusses its requirements. Section \ref{sec:design} outlines the high-level design of our the communication framework and details its internal architecture and Section \ref{sec:evaluation} illustrates its use in two real-world test-beds.
Finally, Section \ref{sec:concl} summarizes the contributions of this paper and
discusses our plans for future work.

\input{relatedwork}

\input{requirements}

\input{design}

\input{evaluation}

\section{Conclusion}
\label{sec:concl}

In this paper, we have described a WSN Communication
Layer that is purposefully built to facilitate the
construction of robotic ecologies.

We discussed a number of requirements characterizing this type of systems and
illustrate how they have been addressed in the design of the
new communication framework. 
The challenge was to develop and satisfy the following conflicting requirements: to
provide homogeneous communication services and, at the same time, to integrate devices with very
different communication and processing capacities. To this end, the approach
chosen in \prog was to build communication services on top of
established robotic and WSN middlewares.

In particular, the framework has been
purposefully designed to be as open, flexible and extensible as possible. For instance, in the AAL test-bed the Java proxy
made the data gathered by the sensors available to the higher layers without the need to account for specific network protocols. Concerning the interoperability requirement,
the Hospital scenario demonstrated the coexistence of several different standards of communications, languages, and environments, such as TinyOs, Java, Konnex, PEIS. TCP-IP.
In both scenarios, we have also seen the support for dynamic systems where the devices joining and leaving the system at run-time on completely different topology deployments.
The proxy also shown how we have fulfilled the Configuration and Control requirement, which handles PEIS subscriptions to data gathered from the WSN by activating and
configuring the sampling of sensor data.
 
The software of the Communication Layer is available for download
under an Open Source license at \texttt{https://github.com/gunther82/RUBICON-CommLayer}.

%
%


\bibliographystyle{plain}
\bibliography{rubicon}

\end{document}

%% file: relatedwork.tex
\section{Related Work}
\label{sec:relwork}
\noindent
%

There are many examples of related work combining wireless sensor networks (WSN) with
mobile robots. WSNs are usually used to
report events that need further investigation and intervention
by the robots. In some cases, the robot can also extend the monitoring capabilities of WSNs \cite{5876332}.

On the other hand, robots mobility helps
the WSN to monitor and operate in a larger area than is
possible with fixed sensor deployments. 

Mohammad Rahimi et al. \cite{1241567} studied the feasibility
of extending the lifetime of a wireless sensor network by
exploiting mobile robots that move in search of energy,
recharge, and deliver energy to immobile, energy-depleted
nodes.The PlantCare project
\cite{springerlink:10.1007/3-540-45866-2_13} has demonstrated
how a robot can be used to deploy and calibrate sensors, to detect
and react to sensor failure, to deliver power to sensors, and
otherwise to maintain the overall health of the wireless sensor
network. 

Navigation strategies employing WSNs usually rely on
the fact that the positions of all network nodes are well known
or can be inferred. The solutions for the localization problem
often employ RSSi readings, which are well documented as
unreliable in dynamic environments, to determine node or robot
positions \cite{5152449}, 
often as part of Robot SLAM (Simultaneous
Localization and Mapping) solutions \cite{5971336}.

Batalin \cite{Batalin:2005:SCA:1104515} addresses the problem of
monitoring spatiotemporal phenomena at high fidelity in an
unknown, unstructured, dynamic environment. The robot explores
the environment, and based on certain local criteria, drops a
node into the environment, from time to time. Sensor nodes act
as signposts for the robot to follow, thus obviating the need
for a map or localization on the part of the robot.

All these solutions for the integration of WSN and mobile
robotics usually are developed to solve specific problems in
specific scenarios. However, a number of research initiatives
have tackled the creation of generic communication frameworks
to be used within the robot/WSN application domain.

Gil et al. \cite{gil_robotica07} describe a data-centric middleware for
wireless sensor networks in the scope of the European project
AWARE. The middleware implements a high-level abstraction for
integration of WSNs with mobile robots. This is achieved by
providing data-centric access to the information gathered by
the WSN, which includes mobile robotic
nodes. Nodes in the network organize themselves to retrieve the
information needed by the robots while minimizing the number of
transmitted packets in order to save energy. Robots are
connected via a high-bandwidth IEEE 802.11 WiFi network and
interact with the low-bandwidth IEEE 802.15.4 WSN via a Gateway.
The Gateway is in turn connected to both networks and used to collect the data
gathered within the WSN.

The work of \cite{faultIsolNN} focuses in deploying mobile robots
on environments already monitored by unstructured WSNs, for
instance, in applications, such as search and rescue, where the
robots must rely solely on this network for control and
communication purposes. To this end, the mobile robots are capable of communicating
with the WSN, which is also able to access and control the robots. 
The resulting
communication framework addresses bandwidth, message size, and
route restrictions by using adapter components to enable the
communication of robot control messages through the WSN.

Pav{\'o}n-Pulido et at. \cite{pavon2014service} presented an AAL system for monitoring elderly
 people at home, which combines a set of heterogeneous
 components: a service robot, a WSN, a
 smartphone, a tactile tablet, and a personal computer. This work mainly focuses on autonomous
 navigation in the context of AAL.

%
In \cite{4407228}, authors presented a three-layer, service-oriented architecture that accommodated different sensor platforms and exposed their functionality in a uniform way to the business application. This paper focused more on system communication platforms and on interoperability between different firmware systems.

SAL (Sensor Abstraction Layer)  \cite{9998078} is a middleware integration platform designed to provide a consistent and uniform view of sensor networks regardless of the technologies involved. Unfortunately, SAL does not support cheap TinyOS sensors and it is not made publicly available.

Lim et al. \cite{1550845} proposed a novel sensor grid architecture called the Scalable Proxy-based aRchItecture for seNsor Grid (SPRING), which extends the paradigm
of grid computing to the sharing of sensor resources in wireless sensor
networks. The major focus of the study was on the design of the Grid API for sensors.

In \cite{vairo2008secure} the author propose a middleware specifically designed for WSN, developed in TinyOS, with a particular regard for the security and the energy efficiency. It also support the dynamic addition/removal of nodes in the network.

In work \cite{SPE:SPE2114}, Escolar et al. propose an open development framework for WSN applications that can be easily transported to heterogeneous platforms called Sensor Node Open Services Abstraction Layer (SN-OSAL). This approach allows the developers to execute the same application on TinyOS, Contiki, and potentially other WSN OSs. The drawback of this approach is that it forces developers to use a metalanguage of C and then to learn the API proposed in SN-DSL to program portable WSN applications.

In \cite{amato2015querying} the authors address the issue of the mobility in WSN. In particular they present a solution that is able to automatically and dynamically detect and tracking moving events with WSN and to collect useful information from the event while it moves.

To address the issue o hiding hardware heterogeneity in WSNs, middleware approaches are often exploited. Rashid et al. \cite{Rashid2012522} propose a mechanism that enables the communication and interoperation
among highly heterogeneous entities in the form of a middleware, called PEIS, able to integrate robots, tiny devices, and computer--augmented everyday objects into one system.

Regarding smart homes, recent work focuses on the use of middeware to facilitate the integration of various heterogeneous systems \cite{barsocchi2018boosting}.

The RUBICON Communication Layer also builds upon the PEIS middleware \cite{peis} to provide a
decentralized mechanism for collaboration between separate
processes running on separate devices. One of the great advantages of PEIS middleware is that
allows for automatic discovery, dynamic establishment of P2P
networks, self-configuration and high-level collaboration in
robotic ecologies through subscription--based connections and a
tuplespace communication abstraction.

The tuplespace is a shared data space that allows the members of the network to communicate by publishing information as tuples
consisting of a $<$key, data$>$ pair (together with various
pieces of meta-information such as time-stamps, creator, etc.).
This tuplespace
thus constitutes a distributed database into which any
PEIS-component can read and write information regarding any
PEIS-component.

Some of the most popular middleware proposed in the pervasive computing community, like PCom \cite{1276846} and MundoCore \cite{Aitenbichler2007332}, share some characteristics with the PEIS middleware and they are characterized by dynamic reconfiguration and high interoperability. Perhaps the approach closer to PEIS is LIME \cite{Murphy:2006:LCM:1151695.1151698}, which, together with its tailored versions TinyLIME and TeenyLIME \cite{Curino05mobiledata}, is a java-based system for heterogeneous networked devices such as robots and WSN. It is based on a Linda model, where processes communicate through a shared tuple space that acts as the data repository. The LIME systems are interoperable between WSN and larger platforms, including mobile robots.

All these middlewares do not provide explicit support for data streams, which, as we will see, are required for neural computation.

TinyDDS is another middleware which provides interoperable pub/sub solution for WSAN over and TinyOS and Sun Microsystem platforms \cite{boonma2010tinydds}. However, TinyDDS does not consider end-to-end flow control as in our middleware. For the flow control from an access network to a WSN, TinyDDS relies on TCP’s end-to-end flow control.

The communication middleware described in this paper has also been
used to provide a portable demonstration of a cognitive robotic ecology 

Specifically, the system described in \cite{sandygulova2016privet} exploits the plug\&play, peer-to-peer communication between WSN and robots 
and zero-configuration features afforded by the middleware to demonstrate the main concepts behind 
robotic ecologies.
The system shows people a working example 
consisting of robots, sensors and cameras working
together and exchanging information to deliver an interactive demonstration. 
This is able to autonomously
initiate an interaction, present information,
and demonstrate itself to human users,
and is designed to be easily transportable and reconfigurable
in order to adapt to different settings
and requirements.

See \cite{delicato2014middleware} for an overview on existing design approaches for WSN middleware, as well as the most common middleware services and programming abstractions. The paper \cite{curiac2016towards} surveys the current
state-of-art in the field of integrated concept of wireless sensor, actuator and robot networks and presents design requirements and open research issues.

%% file: requirements.tex
\section{Overview of the Communication Layer}
\label{sec:requirements}
%
The main objective of the RUBICON \comm is to provide a communication infrastructure,
on top of which to build cognitively enhanced robotic ecologies.
Such an infrastructure should deal with how data and functionalities are shared among each hardware device and software element in the ecology, and should enable the provision of useful services through the combined use of its heterogeneous components.
In order to draw a set of requirements for the \comm, here
we first give a brief overview of how a robotic ecology is typically organized.

\subsection{A Reference Architecture for Cognitive Robotic Ecologies}
\begin{figure}
\centering
  \includegraphics[width=14cm]{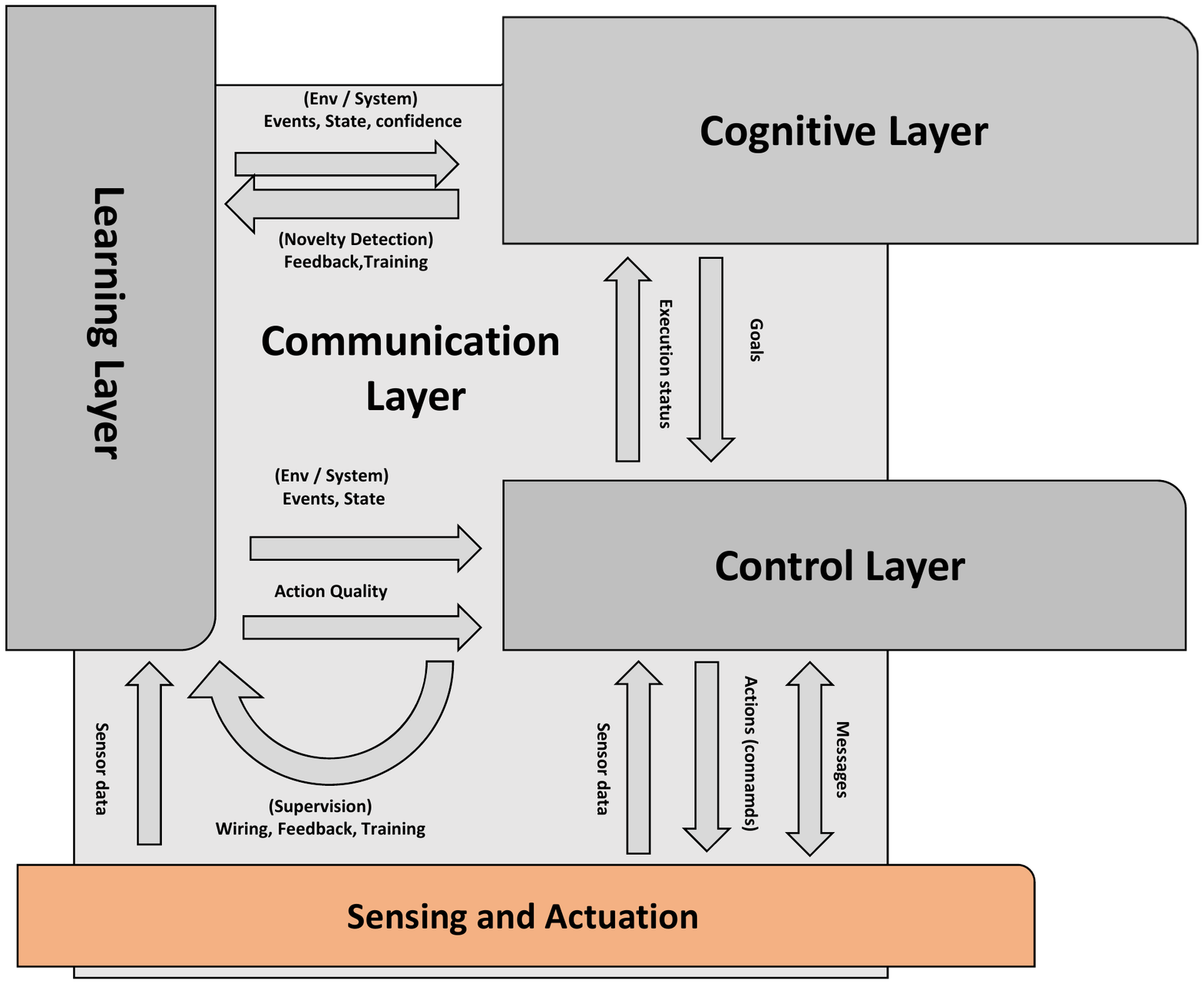}
\caption{Conceptual architecture of the RUBICON system outlining the main interactions of the Communication Layer.}\label{figure:architecture}
\end{figure}

RUBICON has defined a reference architecture \cite{amato2012when}
(see Figure \ref{figure:architecture}) for a robotic ecology hierarchically partitioned into two levels, respectively: (i) the \emph{Coordination Level}, and (ii) the \emph{Service Level}.

The central component at the coordination level is a \emph{\bf Configuration Planner} \cite{di2013robots} in charge of deciding which of the capabilities available in the ecology, in combination, should cooperate to achieve necessary and meaningful tasks, and what information should be exchanged in the process. For instance, in a domestic setting, the planner may decide to send a cleaning robot to the kitchen, and instruct a ceiling camera to help the robot by tracking its location.
One of the key advantages of using planning techniques with robotic ecologies is the possibility
of using alternative means to accomplish application goals when multiple courses of action are available. This increases the robustness of the overall system by exploiting the redundancy of having many devices with overlapping capabilities. For instance, a robot may be able to localise itself through the use of its on-board laser sensor, but also by detecting RFID tags placed under the floor, if the laser-based localization fails for some reason.

The planner may be requested to provide a service either directly by the user, e.g. availing of some user interface, or automatically. In other articles (e.g. \cite{Amato2015}), we have described how the latter can be achieved by exploiting a cognitive reasoning system to reason over past observation of user and system behaviour. For instance, by observing past instances in which the user has often summoned her cleaner robot to the kitchen after eating her meal, the system learns to send the robot in similar situations without waiting for the user's request. 

Components in the service level are logically associated to robots, actuators, appliances, and WSN nodes (also called motes). The control parts in these components are in charge of instantiating and monitoring the action and configuration strategies to be carried out by these devices. Note how the software in these devices are usually built over application-specific frameworks. For instance, the robots in all the RUBICON test-beds are controlled with the Robotic Operating System (ROS) \cite{quigley2009ros} - a component-style framework that facilitates publish-subscribe interaction between the (C/C++) software modules, while WSN nodes are programmed in nesC, component-style embedded version of the C language included in the TinyOS framework\footnote{http://www.tinyos.net/}.

The key to make such a robotic ecology able to adapt to changing and evolving situations is to 
try to make sense of all the sensor data that can be collected with the sensors installed in the environment in order to assess the status of the environment and the application (context). 
This is a normal requirement in smart environments, where numerous other works have harnessed machine learning techniques to learn to recognize user's activities and other context from patterns of noisy sensor data. However, while the majority of existing learning solutions apply a data-centric perspective, in which wireless sensor nodes are used for data collections and feature extraction while the actual learning is performed (usually offline, on more capable computers) learning in a robotic ecology is achieved in a distributed manner.
To illustrate this concept, RUBICON has developed a distributed, adaptable and general purpose memory and learning infrastructure comprising learning neural networks residing on multiple nodes of the robotic ecology. The RUBICON \emph{\bf Learning Layer} can recognize relevant situations out of streams of raw sensor data, and it can process both binary and non-binary sensors to provide short-term predictions based on temporal history of the input signals. For instance, the Learning Layer can be trained to forecast the exact location of the user by examining the history of the radio signal strength index (RSSi) received from a wearable device worn by the user [44], or to provide timely and predictive information on the activities being performed by the user, such as recognizing that the user is eating by analysing the signal and temporal pattern received from sensors installed in the kitchen [50]. 
The Learning Layer is completed with the Learning Network Manager module. This is a Java software agent that offers mechanisms for the configuration and control of the devices participating in the Learning Layer, and which uses the Communication Layer to publish the Learning Layer's outputs to the other upper layers of a RUBICON system.

One of the major strengths of the architecture depicted in Figure \ref{figure:architecture} is the scalability afforded by its hierarchical distribution of information: sensor data is processed as much as possible locally on software components embedded on computational constrained and robotic devices. This reduces the communication overhead imposed on the wireless sensor and actuator network, while the higher levels of the architecture (in charge of planning, cognition and learning management) process the resulting (and compacted) event information and generally interact among each other over the more capable, peer-to-peer (WiFi) network.
Finally, note how such an architecture can be easily mapped to standards emerging from Internet of Things (IoT) initiatives, e.g. in terms of reference architectures and platforms \cite{jin2015etsi}. These commonly exploit a gateway as a forwarding element, enabling various local networks and heterogeneous devices to be connected and engage in lateral collaborations. The gateway in these systems acts as common access point where services can be deployed and executed to configure and monitor the operation of multiple (and simple) devices (what is commonly referred as Edge-intelligence model by the IT community). Such an approach allows to define complex services by exploiting the flexibility and modularity afforded by employing simple devices of limited capabilities and standard interfaces.

\subsection{Communication Requirements}
\label{sec:commrequirements}
We can now draw a number of requirements that the \comm should satisfy in order to enable the collaboration between separate processes running on separate heterogeneous devices, including the high-level cognitive functionalities outlined in the previous section, and fit them with the underlying sensing, acting and learning infrastructure.
Some requirements are more general as they regard functionalities that may also be present in other communication subsystems that address heterogeneous devices, while others are specific to an adaptive robotic ecology as they regard the ability to support distributed neural computation.

More specifically, the generic requirements of the Communication Layer are:

\begin{enumerate}
\item {\textbf{Transparency}}: A robotic ecology consists of a heterogeneous set of devices communicating with each other. Transparency is the overarching requirement for a communication middleware for robotic ecologies. The communications specification should hide the underlying platform differences and decoupling the applications from hardware/OS platforms.
\begin{enumerate}
\item {\textbf{Support for varying computational constraints}}: This is also a primary priority, as target environments will contain devices such as computers with large processing and bandwidth capacities, as well as much simpler devices such as micro-controller-based actuators and sensor nodes with very small memory, computation power, and battery power.
\item \textbf{Sensing and Synaptic Channels} The Communication Layer should allow the other layers to sample and communicate all the sensed data. These include data gathered from sensors installed in the environment (for example switch sensors, occupancy sensors and a variety of analog sensors), information about transmission power (in the form of RSS, the Received Signal Strength indicator) to support localization tasks of the Learning Layer, virtual sensors representing the status of effector (for example to signal the status of appliances) etc.
In particular, learning modules needs to be fed with data streams that are either directly connected to transducers, or that are the result of the neural computation of another learning module. 
\end{enumerate}
\item {\textbf{Interoperability}}: In each device, components may be developed in their own programming language using different APIs. In order to enable composition of applications based on legacy components/networks, and to easily support their replacement and/or update, the software should be interoperable and extensible as possible with applications and networks, targeting both capable hardware, such as Java-enabled devices, home automation network standards and the small 8-bit micro-controllers used in wireless sensor nodes operating over IEEE802.15.4\footnote{http://www.ieee802.org/15/pub/TG4.html}.

\item {\textbf{Support for dynamic systems}}: A robot ecology must choose a combination of hardware and software components whose exchange of information and combined ability to change the state of the environment can achieve the desired goals. This must be supported by a
flexible topology, with routing and communication mechanisms able to connect components (both software and hardware) across different devices (robots, appliances, sensors, actuators). These communication mechanisms should also allow one sharing of data while changing communication path-ways to implement different tasks or in response to changing circumstances. In particular, client applications need updated pictures of all the components available in the system (including all the WSN nodes currently active) and, to this end, they should be informed whenever any node joins (as it becomes operative and connect to the network), or leaves the system (as it gets disconnected, breaks, depletes its battery or simply moves out of network range).

\item \textbf{Configuration and Control}. Client applications must be able to send control instructions to the devices interfaced with sensors and actuators. For instance, the Learning Layer Manager needs to set-up new learning modules, and other existing learning modules, while operating in the defining of their input/output connections with the sensors installed on each particular device the Control Layer needs to communicate new set-points and new output values to every actuator it wishes for. In order to support reliable applications, the delivery of these configuration and control instructions should be guaranteed (acknowledged).





\end{enumerate}

%% file: design.tex
\section{The Structure of the Communication Layer}
\label{sec:design}

The Communication Layer implements a transparent and reliable transfer of data between nodes of the \prog robotic ecology. It provides both connection-oriented and connectionless services and enables the interaction of heterogeneous software components running on PCs, robots and wireless sensor nodes.
The Communication Layer can keep track of the segments and retransmit those that fail. It also provides the acknowledgment of the successful data transmission and sends the next data if no errors occurred.
One of the key-features of the Communication Layer (as discussed in requirement Sensing and Synaptic Channels in Section \ref{sec:commrequirements}) is to provide support for neural computations performed at a higher layer where several sensor nodes are interconnected into a neural network and they need to exchange data (the outputs of the neural computation executed in the mote) at a given frequency.
The Communication Layer also implements a protocol for the dynamic joining of new motes into the network, that is organized in clusters of motes called \textit{islands}. With this protocol the joining mote notifies the coordinator of the island about its capabilities (i.e., kind of mote and list of transducers and actuators it is equipped with), and receives the \prog address that will be used for the future communications.
Finally, the Communication Layer provides services for the management of the radio. For example, to turn off and on the radio interface, for energy consumption purposes.

%


To maintain accuracy, the Communication Layer was developed using a layered approach similar to the TCP/IP stack architecture. It is composed of two levels: the Network level and the Transport level.
The Network level provides mechanisms for addressing and routing packets between nodes, and for the hardware management of the radio interface.
It implements a basic interface for receiving and sending messages and it allows the higher level to abstract from physical details of the communication interface.

The Transport level provides the interprocess communication
between two processes running on different nodes in either
a connection-oriented or a connectionless manner. The connectionless support is used to send commands to a device (a mote, a robot, a PC), or to retrieve single-shot information from a device, like a transducer's reading or the status of an actuator.
The connection-oriented paradigm provides support for the distributed neural network computation performed by the applications.



\subsection{The Network level}
The Network level provides a basic interface for receiving and sending
messages and allows the higher level to abstract from physical details of the
communication interface.

The \texttt{send} command takes in input the address of the destination device to which
 the message will be sent, the pointer to the data buffer to be sent,
the number of bytes of the message and a flag indicating whether the communication will be reliable or not; if set to \texttt{TRUE}, an ack will be generated at destination side for that message.

The \texttt{receive} event is notified at destination side when a message is received. It provides the address of the source device from which the message is received, the pointer to the data buffer received, the number of bytes of the message, the flag indicating whether the ack is required for the message and the sequence number of the received message that is used for the acknowledgment message.

Due to the limitations of TinyOS, if the radio is busy when a packet transmission is requested by an Application level component (for example because another Application level component is sending a message), a failure event will be raised back to the application that had issued the send request. This problem is typically left to the applications to handle, which have to manage a timer and try later to send the message once the radio becomes available again. In our Communication Layer, this management is no longer required. In fact, the Network level implements a circular buffer that is used to store temporarily and transparently outgoing messages received from the applications if the radio is not available. A similar buffer is used at reception side to temporarily queue all messages received from the network in order to give the applications the time needed to manage the message.

In addition, the TinyOS implementation of the network level provides  functionalities for the energy management of the radio and the serial interfaces. In particular, it implements commands to turn on and off both the radio and the serial interfaces, and commands to get the status of those interfaces.



\subsubsection{The Topology and the Routing of the Ecology.}
\begin{figure}
 \centering
  $\vcenter{\hbox{\includegraphics[width=9cm]{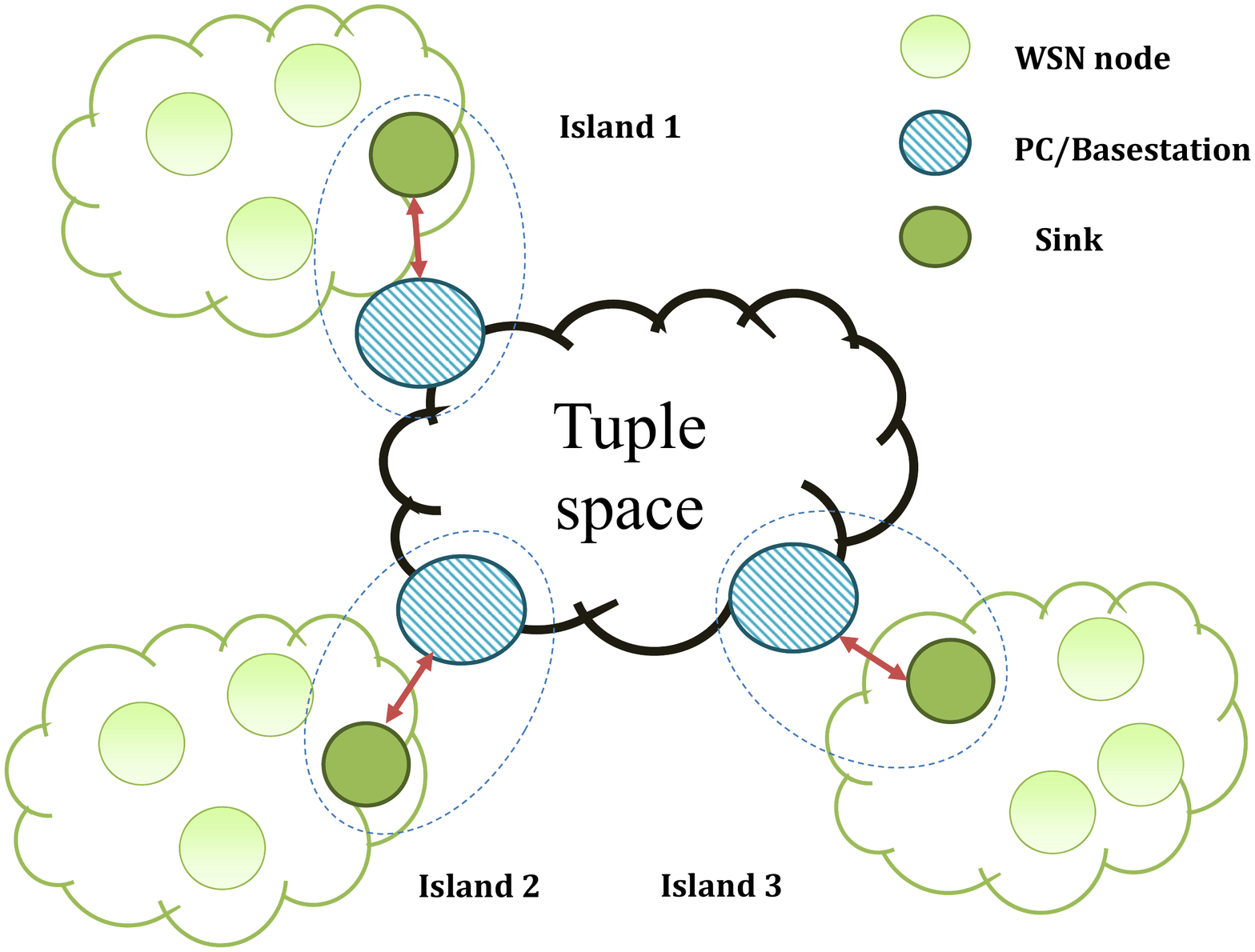}}}$
  $\vcenter{\hbox{\includegraphics[width=4cm]{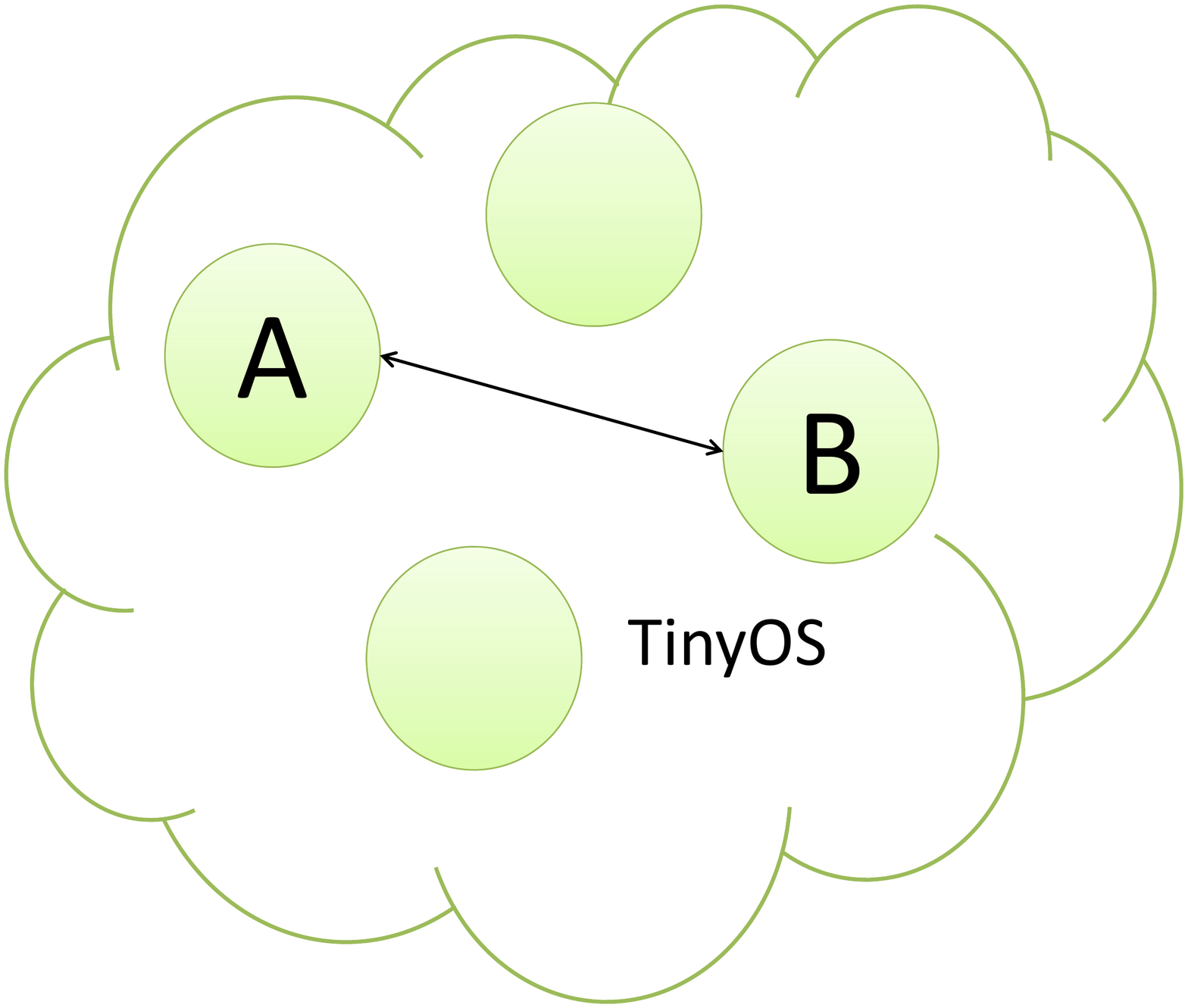}}}$
\caption{The general organization of the WSN network (left). The intra-island communication (right).}\label{figure:WSAN_Architecture}
\end{figure}

The Communication Layer is based on a hybrid topology, in which islands are interconnected by mean of the PEIS ``whiteboard'' \cite{peis} through a special mote acting as a hub. This mote is called \emph{sink} and it is, in turn, connected via USB with a basestation (a PC) that eventually is linked to the PEIS network. The nodes belonging to an island communicate with each other through the shared standard TinyOS channel (see Figure \ref{figure:WSAN_Architecture}, left frame). This type of connection can be seen as a \emph{bus topology}, in which messages are broadcasted on the same communication mean to all nodes. Each node checks the destination address in the message header and processes the messages addressed to it or sent in broadcast. The connectivity between distinct islands relies on a \emph{second level} of connection provided by PEIS. The central hub is crucial to ensure the routing. All the nodes are connected to the central hub (i.e., the sink node together with the basestation) and hence all the messages go through the central hub before going to the individual nodes. The main advantage of this two-layer topology, which resemble a tree topology, is that it is easy to add new nodes into the existing network. In this type of topology, which is often referred in literature as \emph{Cluster-Based Model} \cite{singh2010survey,Abbasi20072826}, the routing is recognized as energy efficient compared with direct routing and multi-hop routing.

In addition to supporting network scalability, the cluster-based model has numerous advantages. It can localize the route set up within the cluster and thus reduce the size of the routing table stored at the individual node, and it can also conserve communication bandwidth since it limits the scope of inter-island interactions to sink node and avoids redundant exchange of messages among sensor nodes.

\subsubsection{The Intra-Island Connectivity.}
The intra-island connectivity is ensured by the communication capabilities offered by TinyOS\footnote{http:\/\/tinyos.stanford.edu\/tinyos-wiki\/index.php\/Mote-mote\_radio\_communication}. Each node of the island is therefore able to communicate directly with other nodes without having to pass from the sink and without the use of algorithms for multi-hop. This has the advantage of simplicity, but requires that each node is within the communication range of the other nodes (see Figure \ref{figure:WSAN_Architecture}, right frame).

\subsubsection{The Inter-Island Connectivity.}
Connectivity between islands is guaranteed by the PEIS middleware. Whenever a node of an island (say island 1) needs to send a message to a node in a remote island (say island 2), it must send the message to the sink node of the island 1, which, in turn, forwards it to the basestation to which it is connected. The basestation of the island 1 then sends the message to the basestation of the island 2 by means of the PEIS middleware. The basestation of island 2, upon receiving the message, forwards it to the sink connected, which, in turn, transmits the received message to the destination node of island 2. 

\begin{figure}
 \centering
  $\vcenter{\hbox{\includegraphics[width=6.5cm]{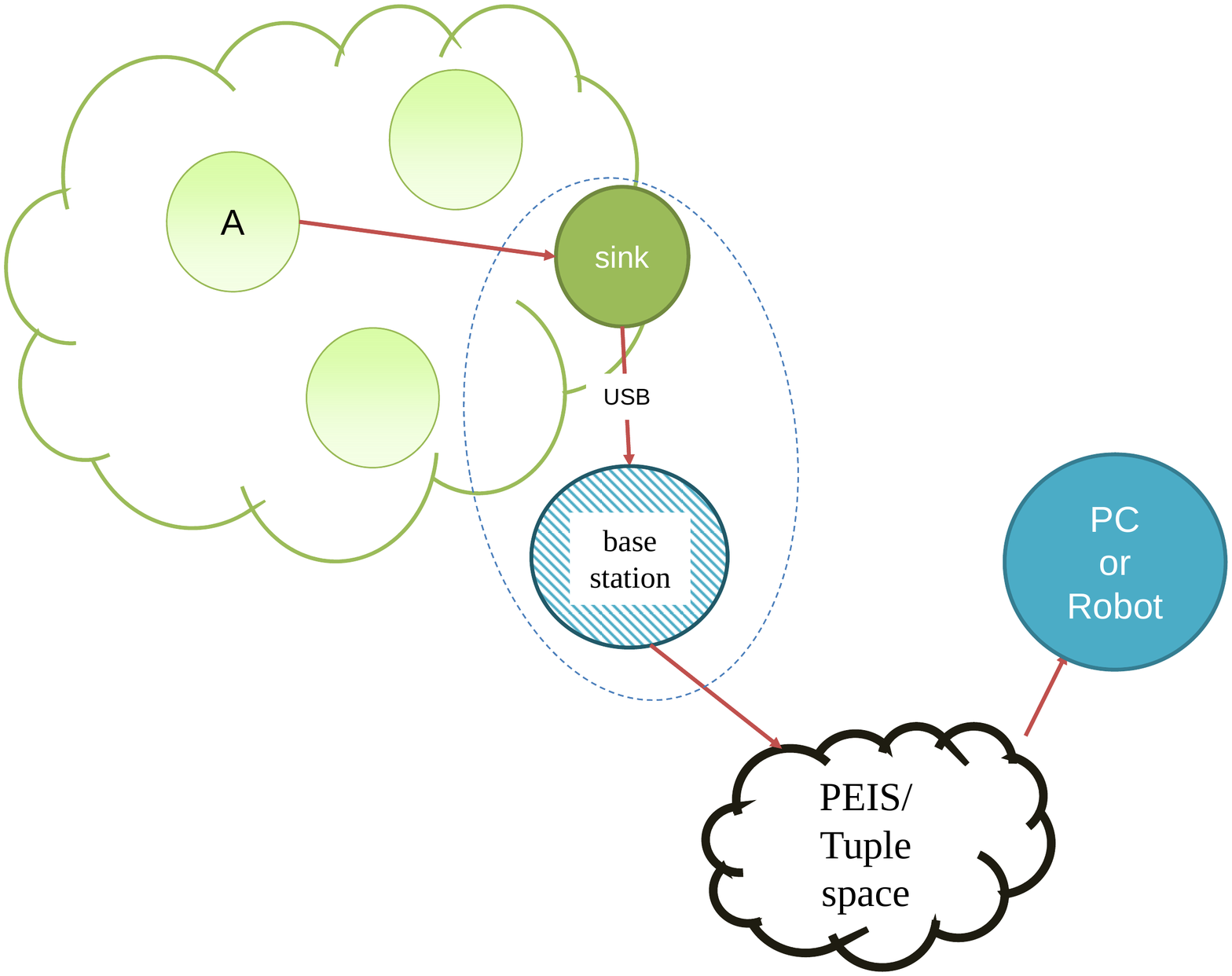}}}$
  $\vcenter{\hbox{\includegraphics[width=7.5cm]{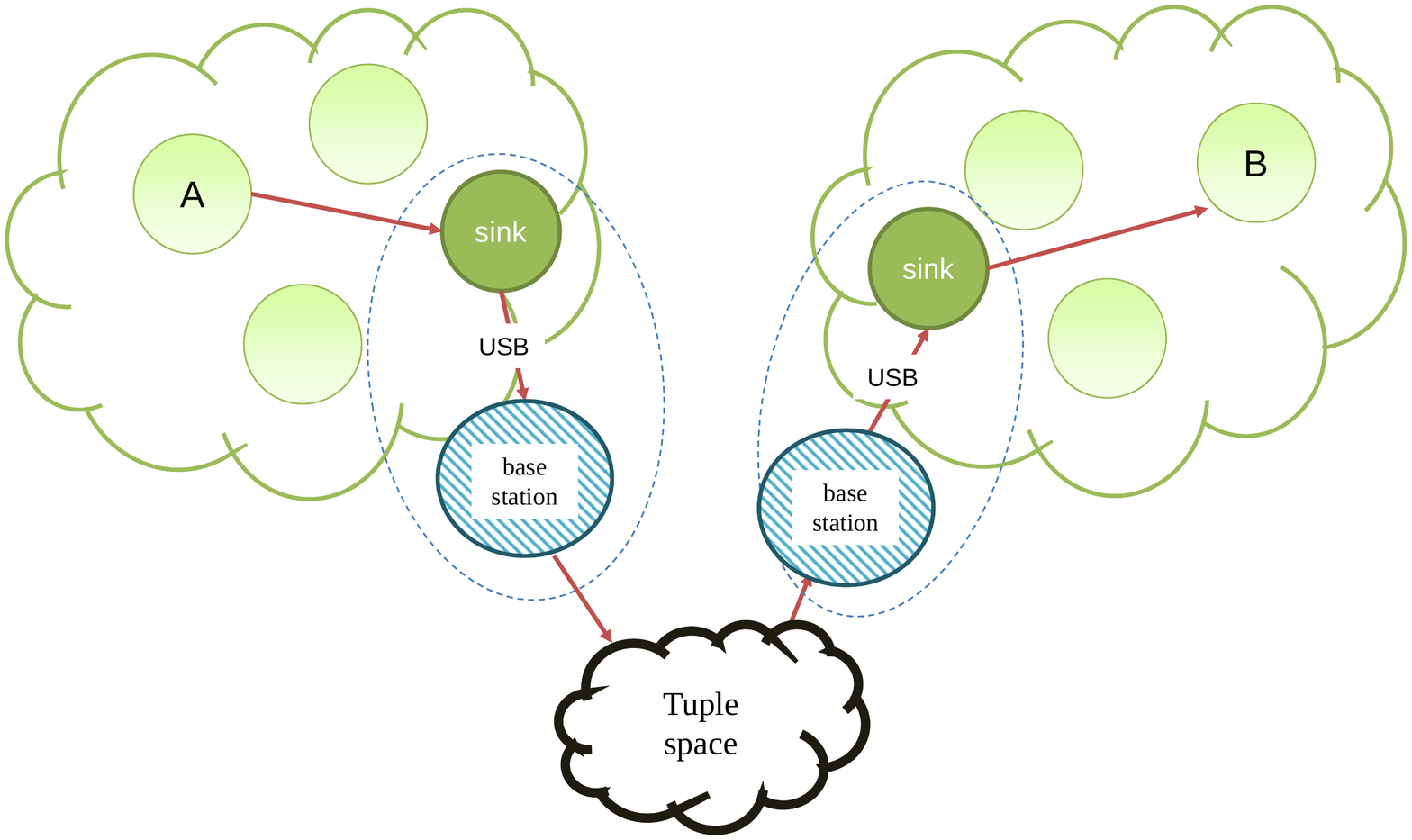}}}$
\caption{The inter-island communication between a mote and a robot/PC (left) and the inter-island communication between two foreign motes (right).}\label{figure:inter-island}
\end{figure}

Figures \ref{figure:inter-island} show the message routing from a mote to PC/Robot connected to the tuple space (left frame), and the message routing from two motes belonging to different a island (right frame). The interaction reported in the figure is bi-directional for both the scenarios. Note that, in \prog a PC/Robot behaves as basestation of an island without motes.

\subsubsection{The RUBICON Addressing Scheme.}
\label{sec:rubiconaddress}
Addressing algorithms in a WSN have to deal with storage consumption costs, processing costs, and the energy since these measures are important in evaluating the performance and improving the scalability of an addressing scheme.
In particular, the design of the \prog addressing scheme must satisfy two conflicting needs, compactness and simplicity, and the ability to target and route messages to elements within the ecology that may be sensors inside the islands or a PC/Robot connected to PEIS.

\textbf{Compactness}. WSN nodes have strong memory constraint. Applying current address such as 32 bit IP addresses or
the 128 bit IPv6 addresses is possible, but results in needless overhead. Significant savings can be realized by using only the minimum number of bits required to uniquely address all the nodes. However, an address-based WSN has its side effects because constructing and maintaining an address structure usually requires additional cost. It is important to consider this cost in WSN, where every bit transmitted can reduce the lifetime of the network. A key factor to consider is that both the packet size and data rate in WSN are usually very small.

\textbf{Simplicity}.
The low processing capabilities of motes could make inefficient the use of multi-hop routing protocols. Moreover, multi-hop routing in WSN is affected by new nodes constantly entering or leaving the network. Our cluster-based routing protocols based on a hierarchical network organization provides an efficient solution and at the same time, it is simple enough to be managed.

\textbf{Dynamicity}.
The \prog address is defined as a pair: \texttt{<pid, devid>}, 
where \texttt{pid} is the PEIS identifier and it addresses autonomous software components running in a PC or a Robot, and \texttt{devid} is the device identifier. The former identifies also the island where the PC is connected, while the latter is used only to address a mote deployed in the island identified by the \texttt{pid}.

The address of the island is dynamic, i.e., it is assigned to the motes at runtime rather than at compile time. In particular, it is received from the sink during the execution of the join protocol, in this way a mote can join any island.
The device address corresponds to the physical TinyOS identifier assigned to the mote while flashing the program in the mote memory.
The separation of the address in island address and device address allows motes deployed in different islands to have the same TinyOS identifier (the device address). In this way, there is no need to know the addresses of the local motes in the other islands when deploying a new island in the ecology since there is no risk of address collision.

\subsubsection{Island Collision Resolution and Safe Join.}

One problem that has been addressed in the implementation of the island paradigm, is the possible overlap of the islands. A mote may be in communication range with two sinks of two different islands and create interferences. One possible solution is to take advantage of the multiple channels of the MAC protocol, which, however, are few (four at most).

The problem is addressed by the network level of the Communication Layer by looking at the island address of the recipient. If the receiver node is a sink and the recipient is a remote island then we check whether the message should be forwarded (routing to a remote island) or should be ignored (collision). In particular, if the island address of the sender mote is different from that of the island which the sink belongs to, then the message is ignored because it is from a mote of another island. If, on the other hand, it is the same island, then the message must be sent to the basestation (which in turn will route the message to the remote island via PEIS). In case the receiver is a generic mote of an island (i.e., different from the sink) and the recipient is a remote island then the message is ignored.

\begin{figure}
 \centering
	\includegraphics[width=14cm]{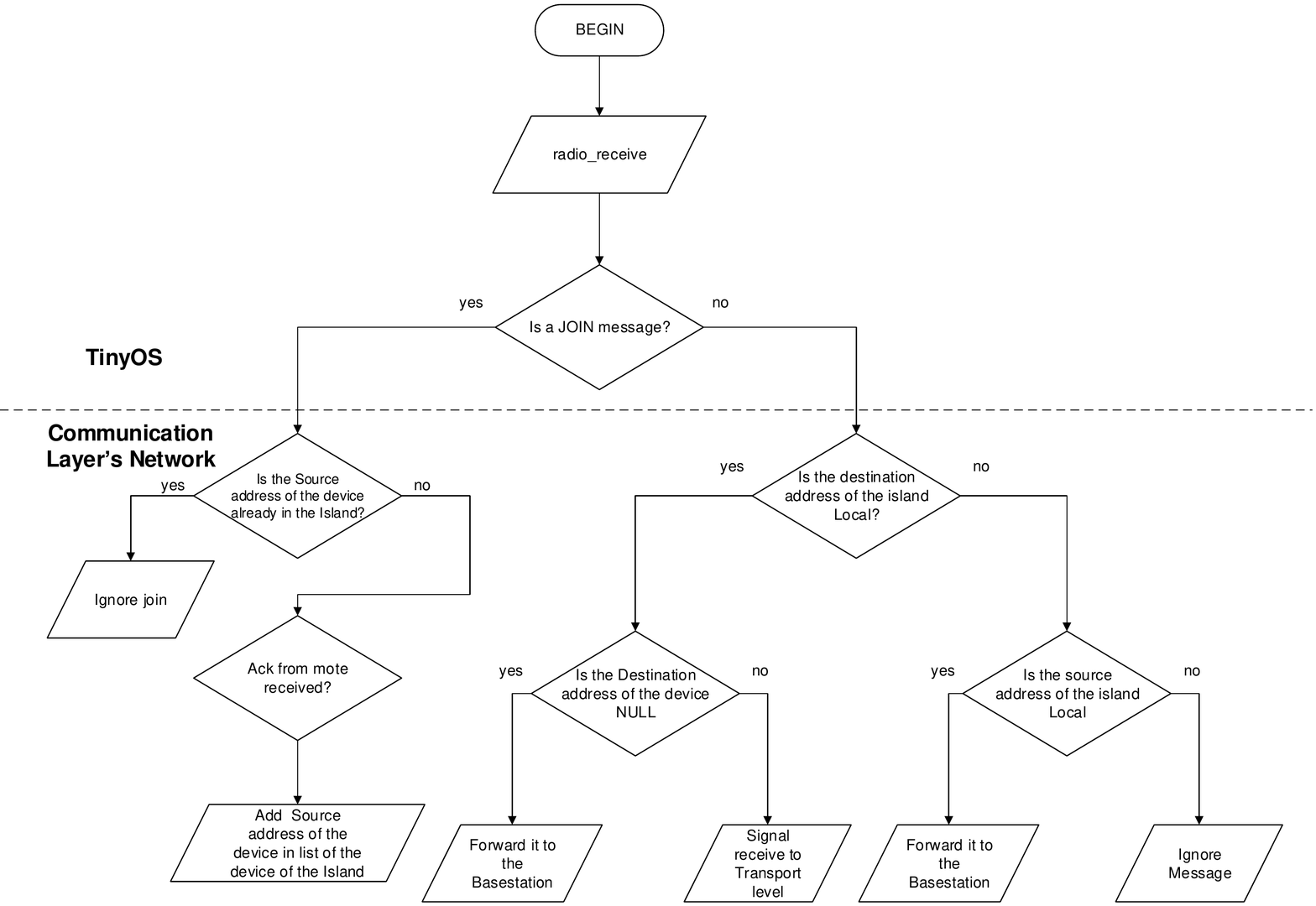}
	\caption{Flowchart of the reception of a message from the sink network level.}\label{figure:flow_radio_receive_sink}
\end{figure}

\begin{figure}
 \centering
	\includegraphics[width=13cm]{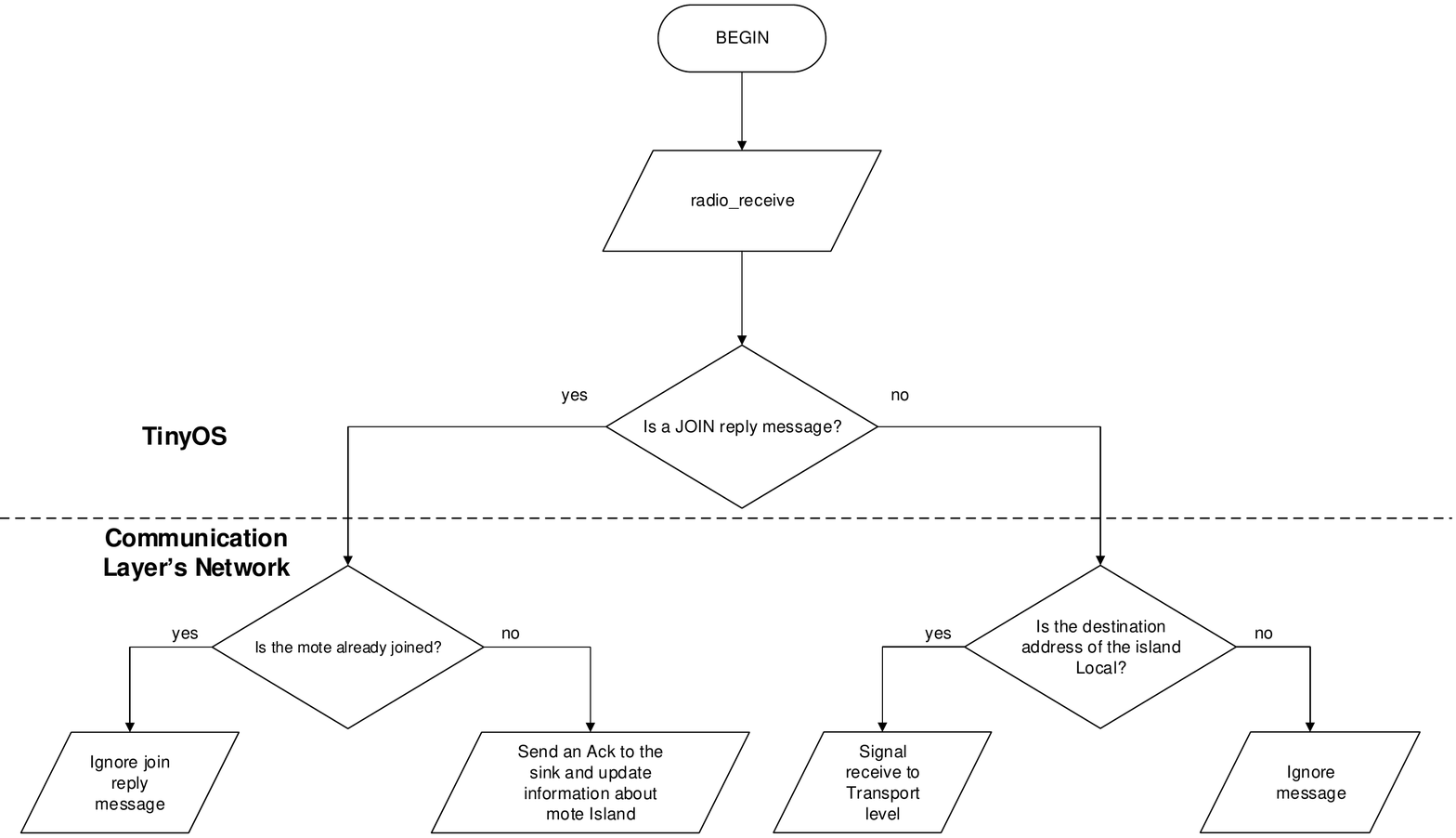}
	\caption{Flowchart of the reception of a message from the mote network level.}\label{figure:flow_radio_receive_mote}
\end{figure}

This behavior is represented by the flowcharts of Figures \ref{figure:flow_radio_receive_sink} and \ref{figure:flow_radio_receive_mote}. The check on the message that avoids island collision is expressed by the right branches of the two flowcharts. The branches on the left side express the join management, which will be addressed in the next section in detail. The sink keeps track of the list of \texttt{devid}s of the motes belonging to its island. When a join message is received from a mote that wants to join the island, the sink checks if the mote is already on the list of the mote of the island. If so, the join message is ignored (because the mote that sent the join message may be trying to join another island and the message has arrived by chance to the sink). Otherwise, the sink sends a join reply message to notify the acceptance of the mote in the island (this message requires acknowledgment). The mote is thus added to the list of the mote of the island, only when the ack message is received by the sink. This additional check is necessary because the join message by the mote could reach two sinks, with the risk of joining two islands simultaneously. In this case, the mote decides which island to join and sends the ack only to the sink of the island that he decided to join. The distinctions between ordinary messages and join messages is guaranteed by the mechanism of the Active Message Interfaces of TinyOS.

\begin{figure}
  \centering
  \includegraphics[width=11.5cm]{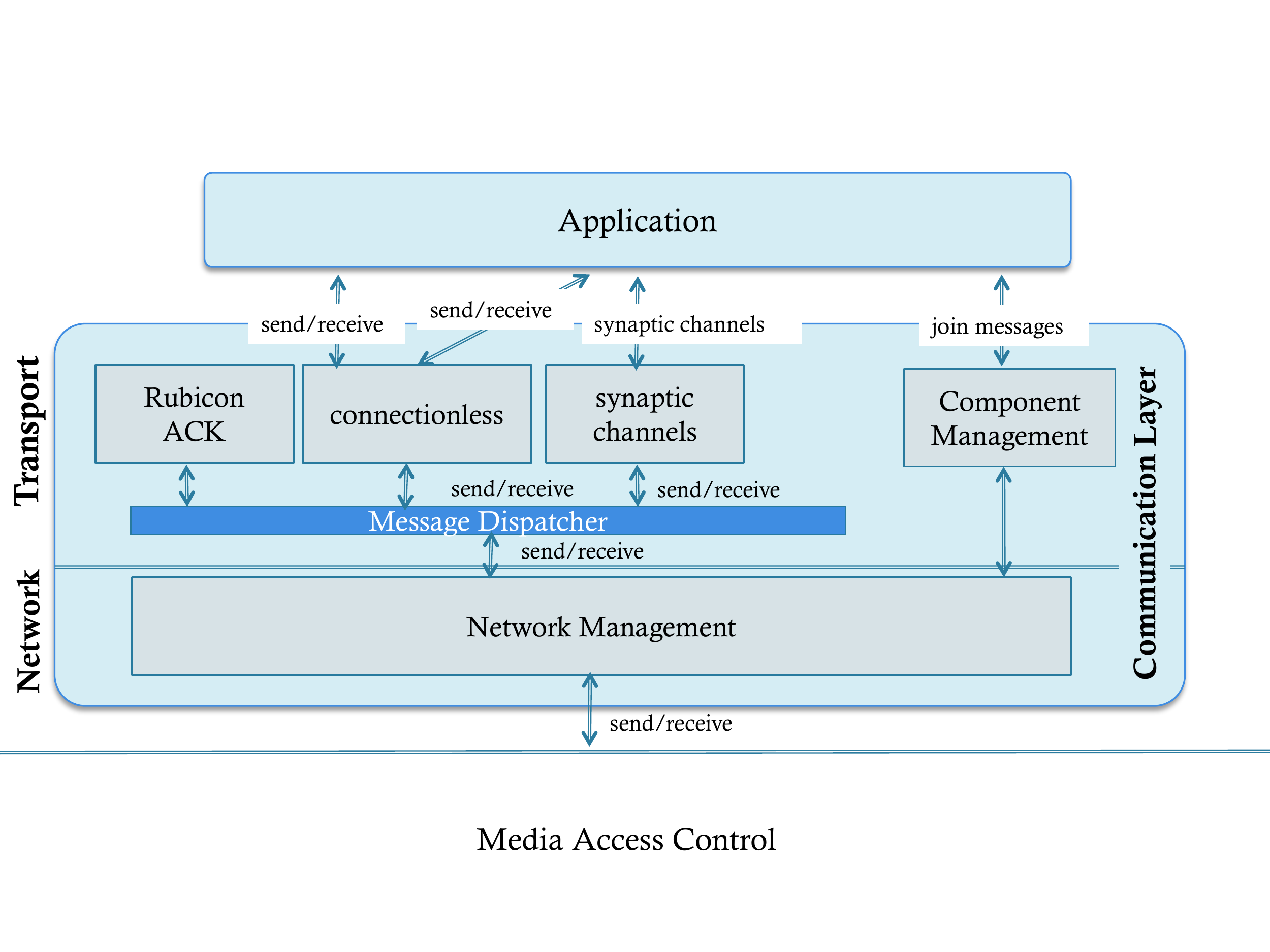}\\
  \caption{The architecture of the Communication Layer}\label{fig:comm_arch}
\end{figure}

\subsection{The Transport level}

The Transport level provides the abstraction of data transport
via the Network level between two processes running on different devices in either
a connection-oriented or a connectionless manner.

In particular, it keeps track of the pending messages (by means of
a sequence number that it is added to each outgoing message),
and provides a reliable communication by implementing an acknowledgment mechanism.

The Transport level also offers a connection-oriented service, the \textit{Synaptic Channel} abstraction, that is used to support the neural computation executed by the Learning Layer. In this case, the Transport level
is  responsible for allocating the data structures needed to
the exchange of data between the devices composing the neural network and to perform the distributed neural computation.
It also implements the join protocol used by motes when entering an island, to obtain the address used for the future communications.
In particular, in order to meet the requirements 1(a) and 1(b) (see Section \ref{sec:commrequirements}), the Transport level provides to the applications the following services:
\begin{itemize}
\item	\emph{Connectionless Message Passing}: this type of communication enables the point-to-point communication between two specific devices of the network, a pc, a robot, or a mote. It is used, for example, for sending commands to remote actuators or robots, or to retrieve a transducer's reading from a remote mote. It can be either reliable or unreliable.

\item \emph{Synaptic Communication}: this type of communication is used exclusively by the Learning Layer and enables two ESNs (Echo State Networks), running on different nodes, to exchange data. In particular it enables the transmission of the output of a set of neurons from a source ESN to a destination ESN.
\end{itemize}

The Transport level is composed of the following software components (see Figure \ref{fig:comm_arch}):
\begin{itemize}
 \item	\emph{Connectionless}: provides connectionless communications between applications of the RUBICON system.

\item \emph{Synaptic Channels}: implements the support for the Synaptic Channel communication paradigm.

 \item \emph{Message Dispatcher}: it is an internal component of the Transport level, responsible to forward a message received from a remote device, to the proper upper-level component. This is done by means of two dedicated fields that are added in the header of each message sent through the Communication Layer, namely \texttt{APPID} and \texttt{TRANSID}. These fields will be discussed in Section \ref{sec:messagedispatcher}.

\item \emph{Rubicon ACK}: it is responsible for generating the acknowledgment when a reliable message is received at Network level.

\item \emph{Component Management}: it is available only in the TinyOS implementation of the Communication Layer and implements the join protocol at both mote and sink side. To this purpose, it receives the join request from the applications and strictly interacts with the Network level to communicate with the sink in order to receive the address to be used in the future communications inside the island.
\end{itemize}

\subsubsection{Connectionless Message Passing.}
\label{sec:connlessmsgpass}

The Connectionless component is used to send commands to the motes, actuators, robots and to receive responses to previous requests. To this aim, it provides a pair of primitives \texttt{send/receive} for exchanging single messages between two devices with or without receipt acknowledgement.


The \texttt{send} command sends a message to the destination node identified by the \prog address (as defined in Section \ref{sec:rubiconaddress}) passed as parameter. It takes in input also the following parameters: the pointer to a memory region where the data to be sent are stored, the number of bytes to be sent, a boolean that specifies whether the communication requires an acknowledgement or not, and  the identifier of the Application level component originating the message (\texttt{APPID}, see Section \ref{sec:messagedispatcher} for details).
The Connectionless component encapsulates this field at the beginning of the payload that is passed to the lower level.
It is used at destination side to dispatch the message to the proper Application level component.

From the receiver point of view, the Connectionless component provides the \texttt{receive} event, which signals that a new message has been received.
It contains the following parameters: the \prog address of the source of the message, the pointer to the received message and the number of bytes received.


\subsubsection{Synaptic Communication.}
\label{section:synaptic}
The learning applications need to communicate through appropriate mechanisms implemented by the Communication Layer. This can be achieved through the abstraction of the \emph{Synaptic Channel}, which provides a point-to-point communication channel between two nodes of the ecology. The synaptic communication is organized in duty cycles of fixed duration that are defined by a clock whose value is configured at runtime. In each cycle the learning network at Application level computes the output of its internal computation (using the input received from remote motes at the previous cycle) and writes them in the output buffer. When the clock fires, the outputs of the learning computation are sent to the connected motes.

On the top of the communication mechanism provided by the Synaptic Channels, the learning applications build the Synaptic Connections, an abstraction of a connection between two neurons residing on different nodes of the RUBICON ecology.

The Communication Layer provides two communication modalities:
\begin{itemize}
  \item \emph{Reliable}: in this modality, the source node always transmits the readings to the remote node at each clock tick. At the destination node, the Communication Layer informs the Learning Layer of any problems with the delivery of the remote neuron readings. For instance, the unavailability of certain neuron readings is signaled by the \comm to the Learning applications so as to allow the \learn to trigger the appropriate fault handlers.
  \item \emph{Power save}: in this modality, the reading of a remote node is transmitted, by the source node, only if the reading has changed since the last clock tick. Unchanged readings will not be transmitted and missing data at the destination node will be interpreted as unchanged data.
\end{itemize}

Note that, the power save modality is transparent to the Learning applications: an unchanged reading that is not transmitted by the source node should anyway appear in the input interface of the destination node and its availability signaled to the Learning Layer. It is the responsibility of the Communication Layer to notify the Learning Layer at the destination node of the availability of the remote input at the clock tick (even if this input is unchanged since last tick).

In the rest of this section we give some examples of use of the synaptic channels. The complete list of the Synaptic Channels API is shown in Figure \ref{fig:synchaninterface}.

\begin{figure}
	\begin{tabular}{| l |}
		\hline
		\texttt{command create\_syn\_channel\_out(dest, size, *params);}\\
		\texttt{command create\_syn\_channel\_in(src, size, *params);}\\
		\texttt{command dispose\_syn\_channel(channel\_id);}\\
		\texttt{command start\_syn\_channel(channel\_id);}\\
		\texttt{command stop\_syn\_channel(channel\_id);}\\
		\texttt{command start\_all\_syn\_channels();}\\
		\texttt{command stop\_all\_syn\_channels();}\\	
                \texttt{command write\_outputValue(channel\_id, pos, output\_data);}\\
               \texttt{command read\_inputValue(channel\_id, pos);}\\
               \texttt{event synData\_received(channel\_id, status);}\\
               \texttt{command setTimeStep(clock);}\\
			\hline
	\end{tabular}
	\caption{Synaptic Channel Interface.}
  	\label{fig:synchaninterface}
\end{figure}

\begin{figure}
  \centering
  \includegraphics[width=15cm]{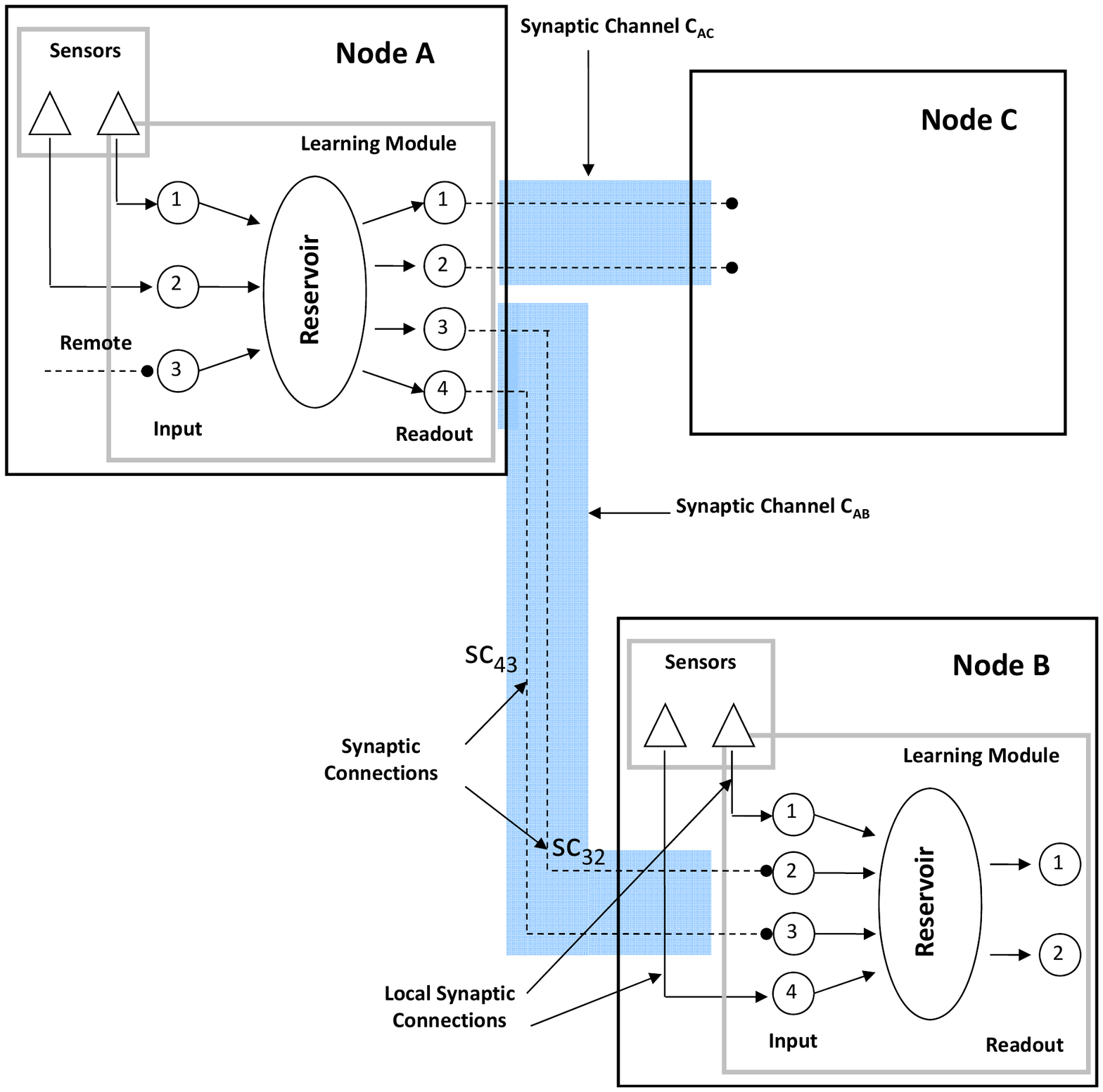}\\
  \caption{Synaptic communications}\label{fig:synaptic}
\end{figure}

A synaptic channel needs to be created at both source and destination sides by means of, respectively, the \texttt{create\_syn\_channel\_out} and the \texttt{create\_syn\_channel\_in} commands. These two commands differ only in the semantic of the first parameter: in the first case it represents the address of the destination end of the synaptic channel, in the second case it is the address of the source end. The other two parameters are: the size of the synaptic channel in terms of synaptic connections contained in the synaptic channel (each synaptic connection corresponds to one output of the learning network, see Figure \ref{fig:synaptic}) and some additional parameters like the modality of the synaptic channels transmission (reliable or power safe).
This command returns the identifier of the synaptic channel created.
The command \texttt{dispose\_syn\_channel} is used to delete a created synaptic channel. It takes in input the identifier of the synaptic channel to be deleted.

%
%


In order to initiate the execution of the synaptic computation, that is sending the output of the learning network to the remote devices, the synaptic channels must be started. It is possible to start a single synaptic channel by means of the command \texttt{start\_syn\_channel}, that takes in input the identifier of the channel to be started, or it is possible to start all the synaptic channels at once by means of the command \texttt{start\_all\_syn\_channels}. It is possible to stop one channel at a time, or all the channels at once by means of, respectively, the commands \texttt{stop\_syn\_channel} and \texttt{stop\_all\_syn\_channels}.

Once the synaptic computation is started, at each clock tick the Synaptic Communication component sends the output values in the buffer of each started synaptic channel to its designated destination. In order to fill this buffer, the application needs to invoke the \texttt{write\_outputValue} command that takes in input the identifier of the synaptic channel to be used, the position of the buffer (that corresponds to the position of the synaptic connection inside the synaptic channel), and the actual data to be sent.

At destination side, when the clock tick fires,
the Synaptic Communication component notifies the application about each started synaptic channel with the \texttt{synData\_received} event. Please recall that the synaptic computation requires that a new data is consumed at each clock tick, therefore this event is signaled at each computation cycle, even if no new data has been received.
The \texttt{synData\_received} event contains the identifier of the synaptic channel and the status of the data (i.e., whether the data is new or not). According to the modality of the synaptic channel, reliable or power safe, the notification of old data may indicate an error in the communication in the first case, or unchanged data, in the second case.

The application then, retrieves the data from the synaptic channel by means of the \texttt{read\_inputValue} command that takes in input the identifier of the synaptic channel and the position of the buffer to be read (that corresponds to a synaptic connection output in the source node).

Finally, the Synaptic Communication component provides a command (\texttt{setTimeStep}) to set the clock tick of the synaptic computation.

Figure \ref{fig:synaptic} shows two neural networks executed by learning modules running in two nodes and the synaptic connections within the synaptic channels. The shadows represent the synaptic channels and the lines inside the synaptic connections.


\subsubsection{Message Dispatcher.}
\label{sec:messagedispatcher}
This component is in the middle between the Network level and the upper components of the Transport level and is responsible for routing messages according to the \texttt{TRANSID} (i.e., the identifier of the Transport level component sending the message). 

rrrr
\begin{figure}
  \centering
  \includegraphics[width=7.1cm]{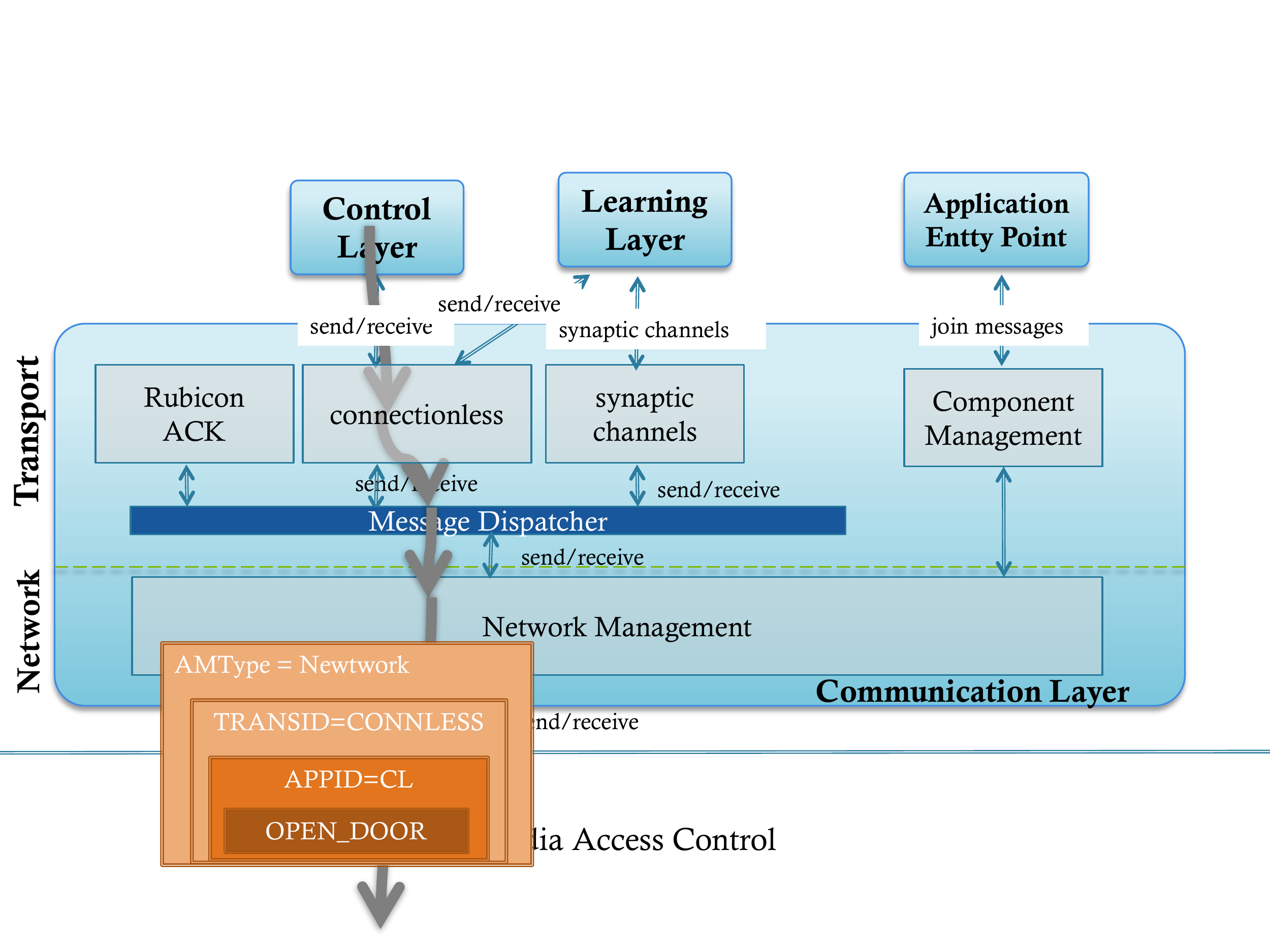}
  \includegraphics[width=7.1cm]{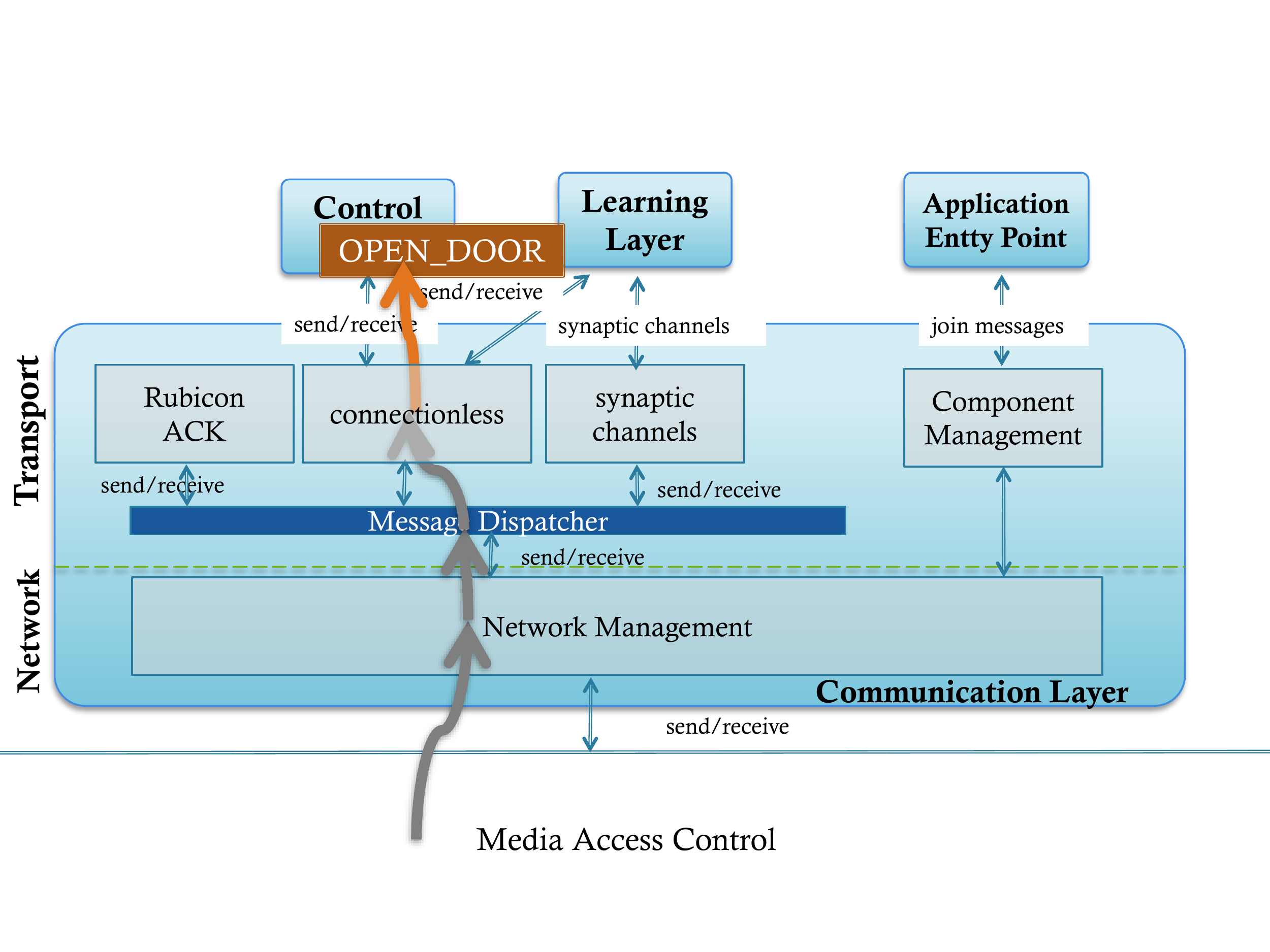}
\caption{The send and receive operations of the Connectionless component of the Communication Layer.}\label{figure:cl_send_receive}
\end{figure}

In order to support flexibility and reusability of applications, we exploit the well-established mechanism of communication called \emph{data encapsulation}, in which a level does not know the details of the information that the upper level wants to send, but the information content becomes a simple payload to be encapsulated and sent to the lower levels along with the necessary information that the level abstraction implements. The opposite path, when the message is received, is characterized by the reverse procedure called \emph{data decapsulation}. We implement this mechanism by means of two fields, \texttt{APPID} and \texttt{TRANSID}, that are added to the header of each message sent through the Communication Layer. In particular, each Transport level component adds the field \texttt{APPID}, identifying the Application level component that originated the message, at the beginning of the message to be sent. Then the Transport level forwards the message to the Message Dispatcher component that adds the field \texttt{TRANSID} identifying the Transport level component forwarding the message, before passing it to the network.
This mechanism increases also the modularity of the Communication Layer. Applications would not have to know the details of the TinyOS APIs, thus more easily supporting further extension of the software and also facilitating its porting to other languages or operating systems.

To this purpose, we use the Active Messages (AM) mechanism of TinyOS. However, this mechanism has the drawback that whenever an application needs to create a new type of message or modify an existing one (for example a control command to an actuator or a learning instruction), we have to create or modify an AM type which implies a modification of the Communication Layer. For this reason, the AM mechanism is used only internally in the Network, with the advantage of guaranteeing the independence of the application from TinyOS.

In order to better appreciate the benefits of data decapsulation, we show an example of its use in case of Connectionless communications.
In Figures \ref{figure:cl_send_receive}, we depict a schematic representation of the \texttt{send} and \texttt{receive} operations of the Transport level. Suppose an application wants to issue a command to a remote mote equipped with actuator capable of opening a door. With the proposed paradigm the application can define its own new message, i.e., any sequence of bits that the lower levels will ignore. The message in the example, which we have called \texttt{OPEN\_DOOR} is sent to the Connectionless component that is responsible for sending messages to individual remote devices. This message is encapsulated in a message with the tag \texttt{APPID = CL}, which is to say that the message was generated by an instance of a Control Layer. The message, then, is sent through the Message Dispatcher component, that encapsulates it in a new transport message with the tag \texttt{TRANSID = CONNLESS}, which is to say that the message was generated by the Connectionless component. The Message Dispatcher sends the message to the Network level, which is in charge of creating an active message of type \texttt{AMType = Network} by inserting the transport message in the payload of the network message. Note that it is only at this point that TinyOS Active Message is created, so that if in the future we want to replace the Network level with a new network connected to another type of media, we would need to change only this part of the system, without modifications to the core Communication Layer.

When the message has been delivered to its intended recipient, it goes upwards to the Application level. At each step, the message is subjected to a decapsulation process to get back the original message. All tags \texttt{APPID}, \texttt{TRANSID}, and \texttt{AMType} are exploited to make the internal routing in the receiving node to bring the message to the right application process that must receive it (similar to what happens with TCP/IP ports). For instance, for the Synaptic communications the \texttt{TRANSID=SYN\_CHANNEL} and \texttt{APPID=LEARNING}.

The Message Dispatcher component provides a command for sending a message to the Network level and an event signaled by the Network level when a message is received.
The \texttt{send} command has the same signature as the Connectionless component (see Section \ref{sec:connlessmsgpass}), with the difference that in this case it contains the Transport level component ID (\texttt{TRANSID}) instead of the Application level component ID (\texttt{APPID}).

Upon receiving a message, the \texttt{receive} event is signaled by the Network level with the sequence number of the message and the flag for the acknowledgment passed as parameters. The Message Dispatcher component extracts the Transport level component ID (\texttt{TRANSID}) encapsulated at the beginning of the payload by the Message Dispatcher component at source side, and forwards the received updated payload to the proper Transport level component accordingly.
When the Transport level component (for example the Connectionless component) is notified by means of this event, it checks the reliability flag, and if true, it starts the procedure for sending the acknowledgement message back to the source by means of the Rubicon ACK component (see next section).


\begin{figure}
\centering
  \includegraphics[width=7.1cm]{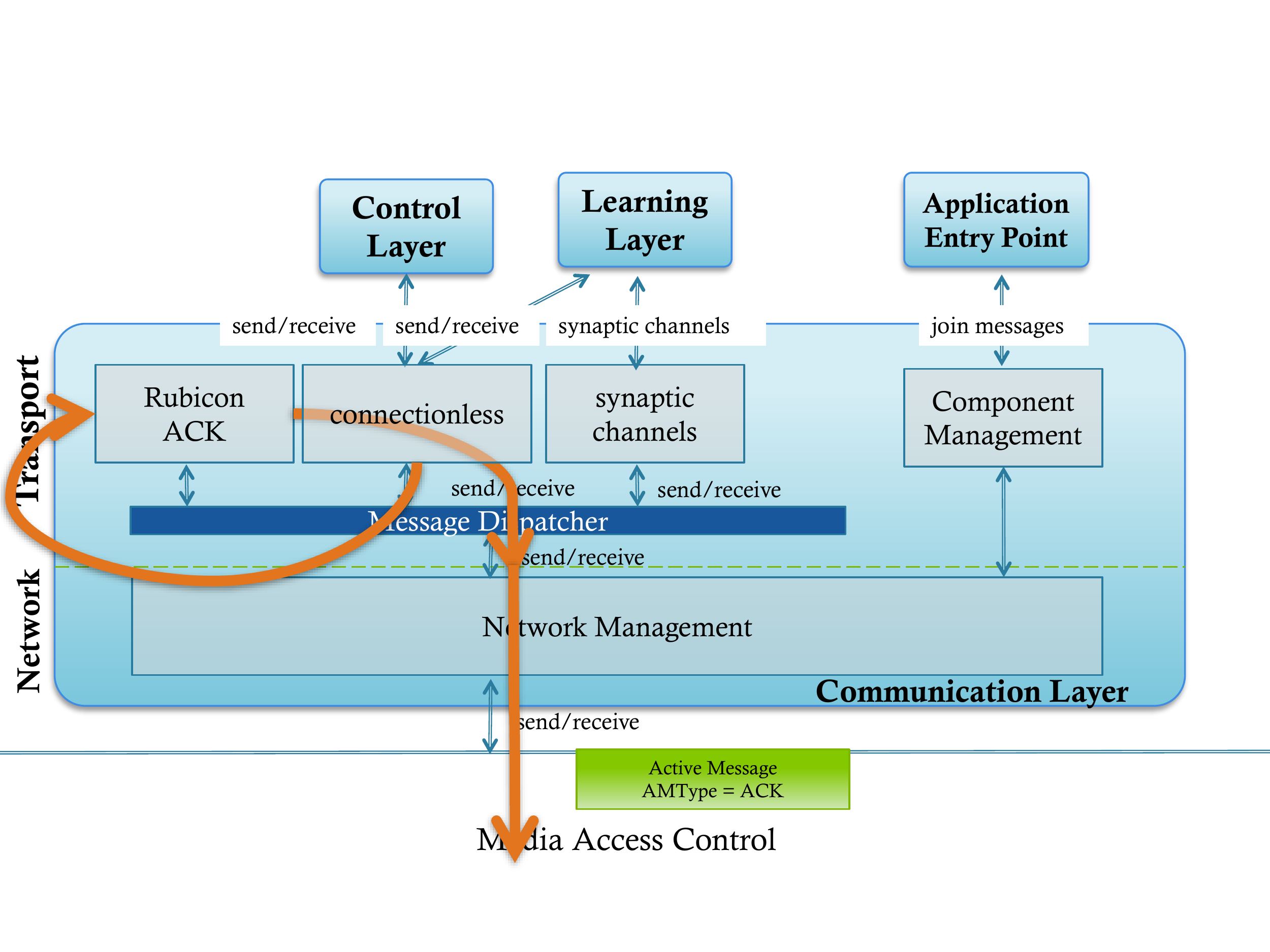}
  \includegraphics[width=7.1cm]{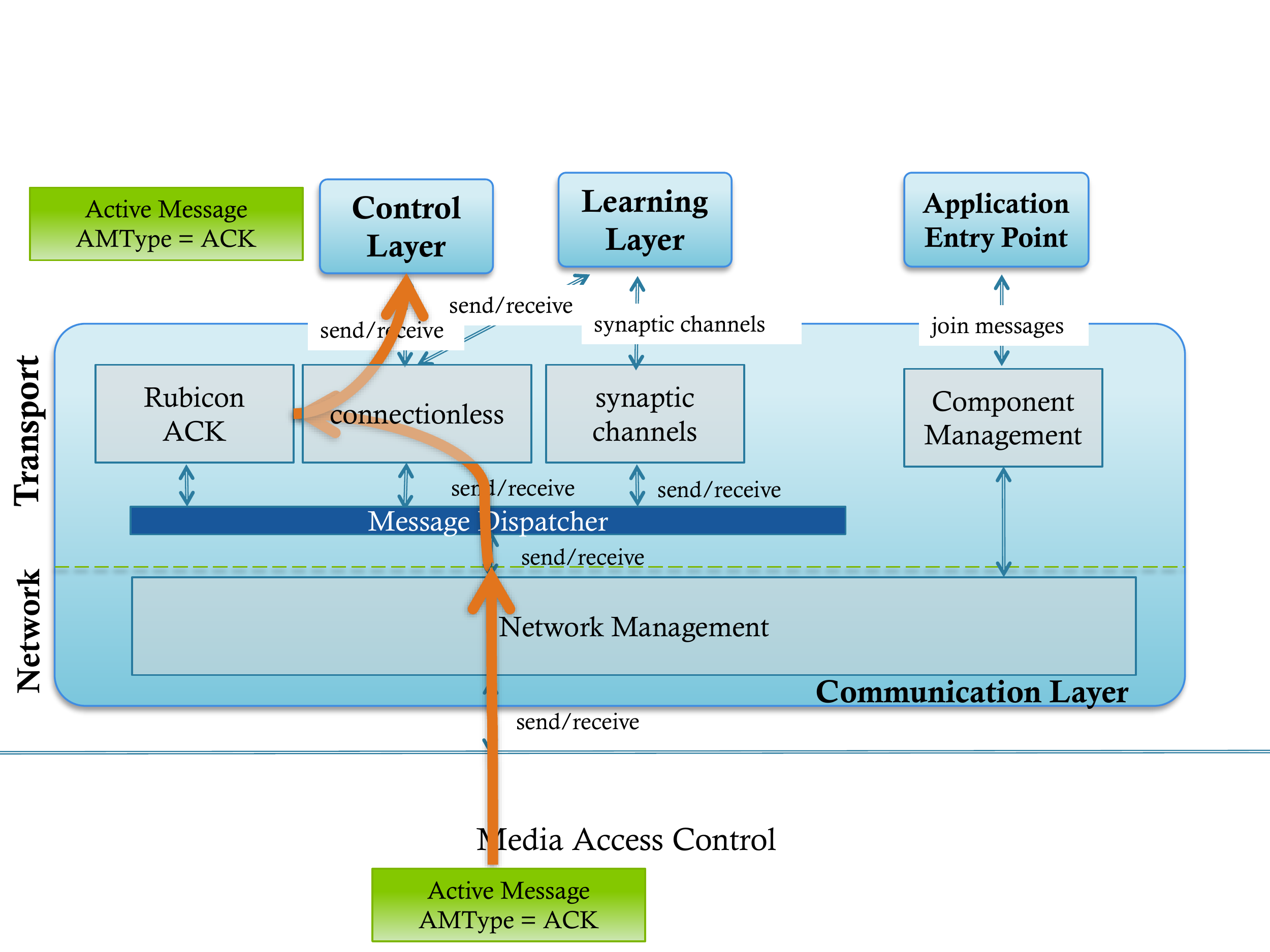}
\caption{The transmission of acknowledgement messages by the Rubicon ACK component introduced in the final version of the Communication Layer.}\label{figure:ack}
\end{figure}

\subsubsection{Rubicon ACK.}
Acknowledgment messages differs from the normal communication messages as they must be processed only by the Transport level that generates them automatically and notify them to the application. If they were generated as normal messages, they should be created and managed by the application as the lower levels do not have access to the contents of the messages. For this reason we have created a specific component, called \emph{Rubicon ACK}, which deals with the generation and the management of the acknowledgement messages in a dedicated protocol.
In particular, it provides the \texttt{send\_ack} command that is used at destination side to generate the acknowledgement message, and the \texttt{receive\_ack} event that is signaled at source side to notify the application that an acknowledgement message has been received.

As anticipated in Section \ref{sec:messagedispatcher}, when a Transport level component (for example Connectionless) is notified of the reception of a message from the Message Dispatcher, it checks the reliability field of the message. If this flag is true, the Connectionless component generates the acknowledgement message, containing only the sequence number received, and passes it to the Rubicon ACK component by means of the \texttt{send\_ack} command. This command takes as parameter the destination address (that is the source of the message to be acknowledged), the payload, the number of bytes sent, the reliability flag (it is always false in this case, we maintained it only for interface compatibility reasons) and the identifier of the application to which the acknowledgement message will be notified (i.e., \texttt{APPID}). The \texttt{APPID} is the same of the source message that is extracted from the payload by the Connectionless component.
The Rubicon ACK component adds this field to the payload of the message and forwards it to the Message Dispatcher, that, in turn, adds the Rubicon ACK component ID (as \texttt{TRANSID} field) to the payload and forwards the message to the Network level to be sent via the physical network.

At source side, upon receiving the message, the Message Dispatcher detects that it is an acknowledgement message through the \texttt{TRANSID} field and forwards it to the Rubicon ACK component. This, extracts the sequence number of the message and the \texttt{APPID}, and notifies the according Application level component about the reception of the acknowledgement by means of the \texttt{receive\_ack} event, specifying whether the acknowledgement was received successfully or not. Figure \ref{figure:ack} depicts this process. When a sender component (in this case the Connectionless) requires an acknowledgement, the receiver component (shown in Figure \ref{figure:ack}, left panel) automatically generates a message with \texttt{AMtype = ACK} and sends it through the network. When the sender receives the acknowledgement (shown in Figure \ref{figure:ack}, right panel) it notifies the application that requested it.

The Connectionless component at source side keeps a timeout for each reliable outgoing message. If a successful acknowledgement for a given message is not received within the corresponding timeout, the \texttt{receive\_ack} event will notify the failure to the Application level component, which, in turn, will decide whether to send the message again or not. 

\begin{figure}
	\begin{tabular}{| l |}
		\hline
		\texttt{command joinIsland(mote\_descr);}\\
		\texttt{event moteJoined(island\_addr, mote\_addr, mote\_descr);}\\
		\texttt{event joined(island\_addr, mote\_addr, success);}\\
		\hline
	\end{tabular}
	\caption{Component Management Interface.}
  	\label{fig:joininterface}
\end{figure}

\subsubsection{Component Management.}
\label{sec:join}
The Component Management is responsible for executing the join protocol when a new mote joins an island. It provides the interface shown in Figure \ref{fig:joininterface}.
When a mote wants to join an island, the application calls the \texttt{joinIsland} command passing, as argument, the descriptor of the mote, which contains information about the type of mote, the list of transducers, and the list of actuators available. For memory occupancy reasons, these lists are represented as bit masks of 16 and 8 bits respectively.

The Component Management then initiates the execution of the join protocol by sending a message containing the mote descriptor received from the application to the sink of the island to join. 
All the messages of the join protocol are transmitted on a dedicated TinyOS communication interface, since the standard communication protocols used in the other transmissions require the \prog address, that will be received by the mote only at the end of the join protocol.
The mote executing the join protocol, also starts a timer upon sending the join request. If the join procedure does not terminate successfully (i.e., all the messages are received with the corresponding acknowledgements) within this timeout, the mote initiates the join protocol again for up four times (this is a system parameter and can be changed).

Upon receiving the join request, the sink stores the descriptor of the joining mote and the sequence number of the request, and sends a reply message containing the \prog address, the current island time used for synchronization and the sequence number received.
As described in Section \ref{sec:rubiconaddress}, the \prog address is composed of the island address and the device address. The island address is simply the PEIS identifier of the island that the mote wants to join; the device address is the same TinyOS identifier assigned to the mote at compile time, and used by the mote in the join request. 

Upon receiving this reply message, the joining mote checks that the sequence number matches the one it used in the request, and if so, it performs the following actions: updates its own address, stores the sink address, synchronizes the current time with the one used in the island and sends an acknowledgement message back to sink. It also notifies the application of the successful joining by signaling the \texttt{joined} event, which contains the address received and the result of the joining.
It also stops the timer of the join. If the timer fires for four times, that is the join protocol failed all the times, the \texttt{joined} event will be signaled with \texttt{FAIL} as result.

\begin{figure}
\centering
  \includegraphics[width=12cm]{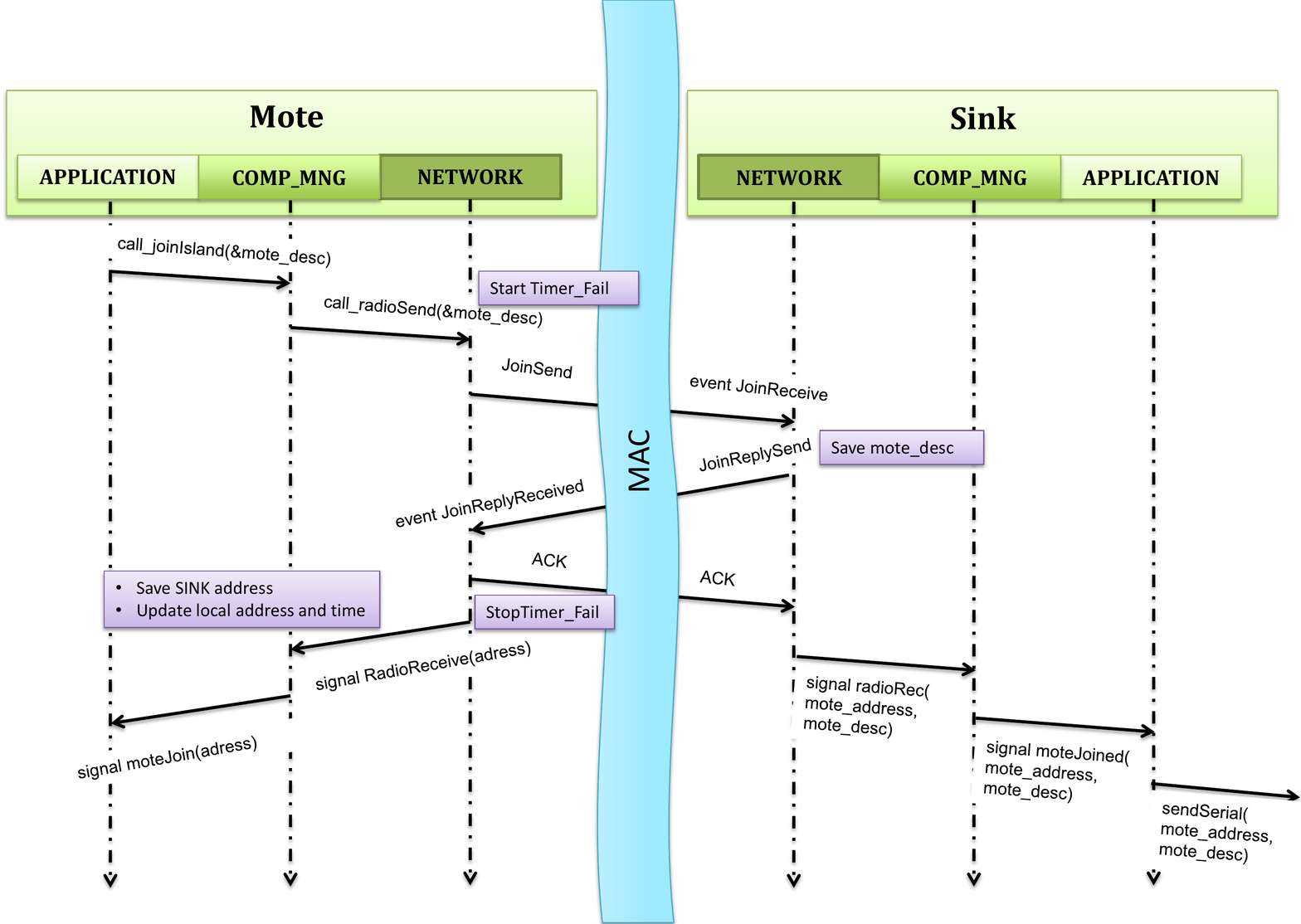}
\caption{Sequence diagram of the join protocol.}\label{figure:ackseqdiagram}
\end{figure}

Finally, when the sink receives the acknowledgement message, it notifies the application about the new mote by signaling the \texttt{moteJoined} event, containing the \prog address of the mote and its descriptor. The application will, in turn, publish this information on the PEIS tuplespace, so that all the devices in the system can interact with the new mote. Figure \ref{figure:ackseqdiagram} shows the sequence diagram of the join protocol.

\begin{table}
  \begin{tabular}{|l|l|}
    
  \hline
    \textbf{Network Layer} & \textbf{1718 bytes} \\ \hline 
    Ingoing radio buffer size & 4          \\ 
    Outgoing radio buffer size & 4          \\ 
    Ingoing serial buffer size & 2         \\ 
    Outgoing serial buffer size & 2          \\
    \textit{Memory for network buffers} & \textit{1464 bytes} \\ \hline 
    \textbf{Transport Layer} & \textbf{644 bytes}\\ \hline 
     Number of input Synaptic Channels & 4       \\ 
          Number of input Synaptic Channels & 4       \\ 
          \textit{Memory for Synaptic Channels support} & \textit{448 bytes} \\ \hline
   \hline
  \end{tabular}
  \caption{Memory consumption of Network and Transport Layers and some system parameters.}
  \label{tbl:memory}
\end{table}

\subsection{Memory Occupation}
The nesC version of the Communication Layer has to face the severe resource constraints of the motes. The  memory of this kind of motes ranges from 8KB up to 10KB of RAM. We developed the Communication Layer by keeping into account this issue and we defined some parameters that can be tuned according to the applications' memory requirements. In particular the network radio and serial buffers and the Synaptic Channels data structures requires a considerable amount of memory. 
The size of each message stored in the buffers is 122 bytes (8 bytes of TinyOS header and 114 bytes of payload). 
The size of the data structure for a single Synaptic Channel is 56 bytes.
We set the capacity of the network buffers (ingoing and outgoing for both radio and serial interfaces) and the number of Synaptic Channels as reported in Table \ref{tbl:memory}. 
The table also reports the memory occupation of these data structures, which constitute the most part of the total memory occupation of the Communication Layer, as can be noted in the Table.

The total amount of memory required by the Communication Layer is 2362 with the parameters configuration reported in Table \ref{tbl:memory}.
These parameters can be tuned according to the application's requirements. In order to compute the available memory for the applications, according to the size of the networks buffers and the Synaptic Channels, we provide the following formula: 

$Comm\_memory = base + b*msg + s*syn\_ch$\\

where, \textit{base} is 450 bytes and includes other variables and data structures of the Communication Layer, \textit{b} is the total number of the positions in the radio and serial buffers (the sum of both ingoing and outgoing buffers), \textit{msg} is the size of a message (122 bytes), \textit{s} is the number of Synaptic Channels allocated, and \textit{syn\_ch} is the size of a Synaptic Channel (56 bytes).

\begin{figure}
\centering
  \includegraphics[width=12cm]{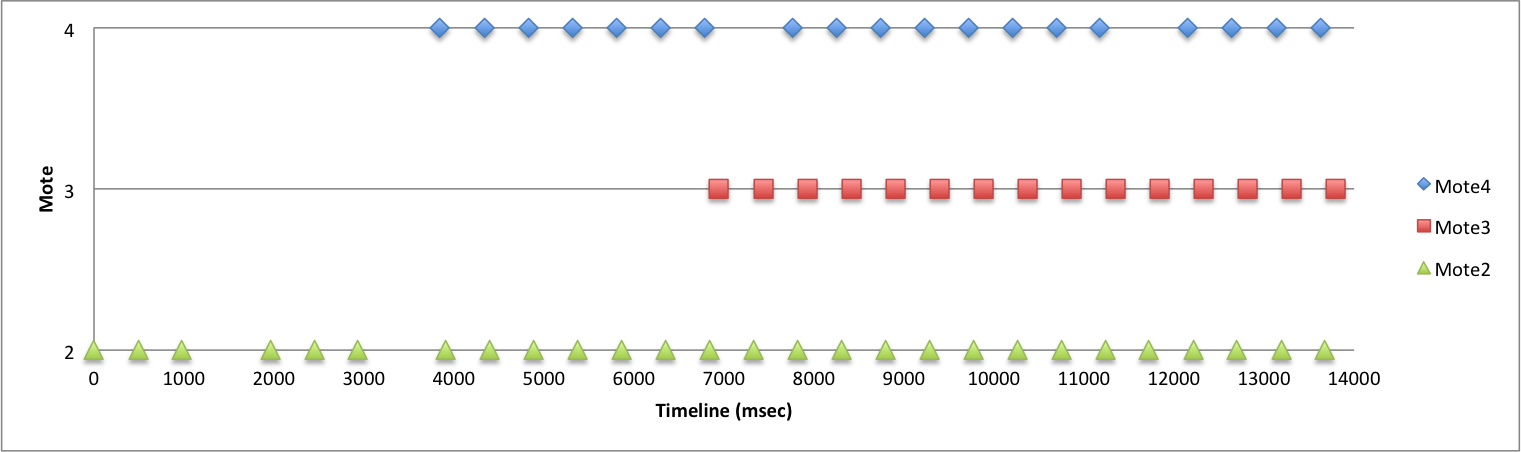}
\caption{Timeline of join and data acquisition from motes.}\label{fig:timeline}
\end{figure}

Figure \ref{fig:timeline} reports an example of execution of a data acquisition task from 3 motes joining the network at different times. Every mote, once joined, start sending its reading to the sink every 500 milliseconds. As can be seen from the Figure, Mote 2 joins the network at time 0, Mote 4 joins at time 3842, and Mote 3 joins at time 6940. In order to make the Figure more readable, these time have been normalized to the time instant of the joining of Mote 2.

%% file: evaluation.tex
\section{Test-Bed Case Studies}\label{sec:evaluation}
In this section, we illustrate the use of the Communication Layer in the two application test-beds and provide some code listing to show examples of synaptic channel usage both in nesC and Java languages.

\begin{figure}
\centering
  \includegraphics[width=14cm]{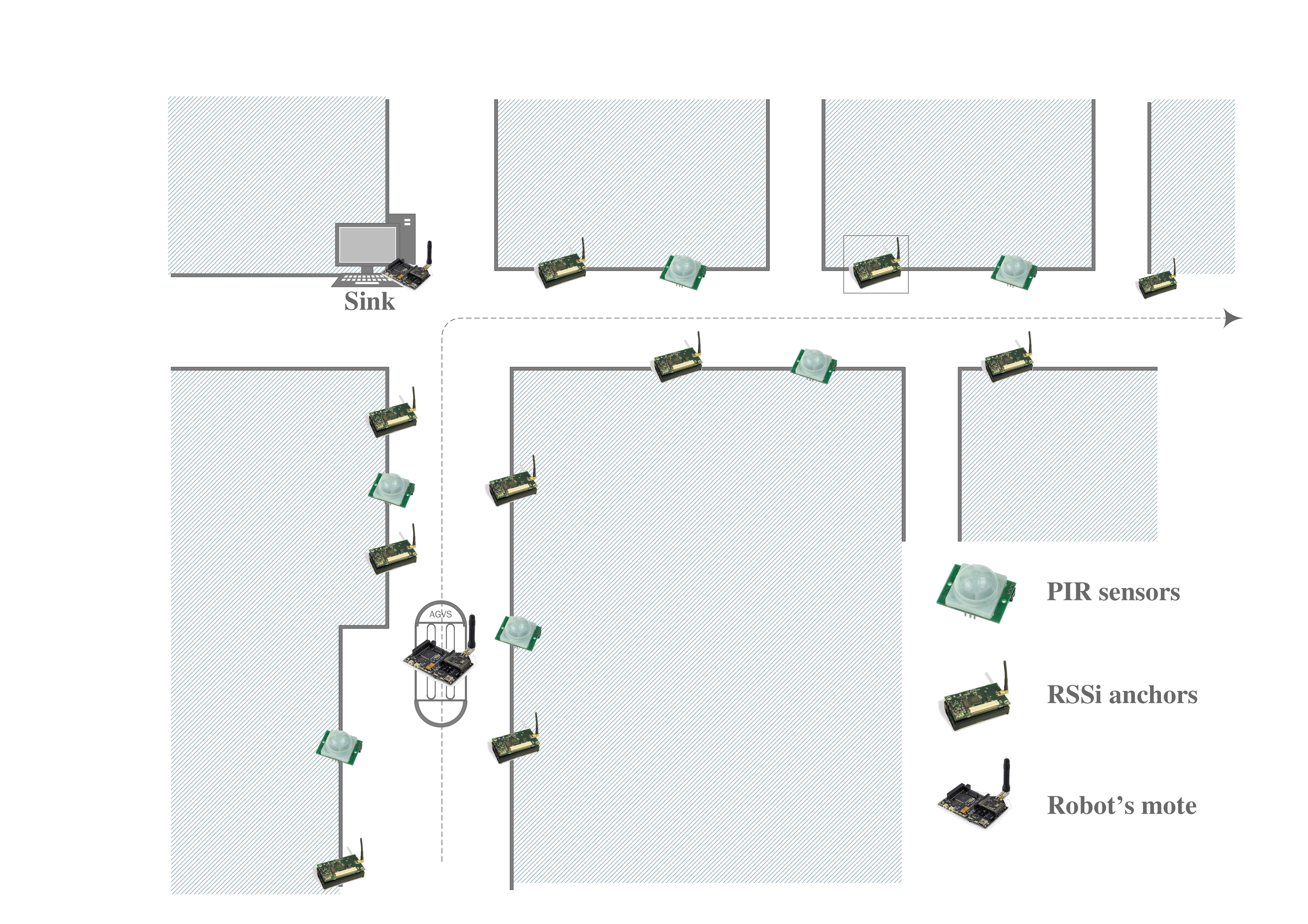}
\caption{Map of the two corridors and AGVS Robot with RSSi telosb mote during the experiments in the hospital.}\label{figure:hospital}
\end{figure}

\subsection{In-Hospital Transport Test-Bed}
The objective of this test-bed is to provide an In-Hospital robot transport system built by combining Robotniks AGVS robots with WSN technology and learning solutions.
A robot localization system based on our Communication Layer was installed in a section of an hospital with one AUGV transport robot. Figure \ref{figure:hospital} illustrates the map of the two joint corridors we used for our experiments as well as the position of the motes in the wireless sensor network. In order to collect RSSi data to exercise the Learning and the Cognitive layers of our RUBICON architecture, we placed ten motes (thereafter called anchors) at equal distance and on alternate sides along the length of the corridors of the hospital, at an approximate eight of 1.5 m. We also placed six motes equipped with PIR (passive infrared) sensors on the ground plane, to detect the movements of the robot, and we attached one mote on the robot. Finally, one sink mote was installed at the junction of the two corridors, in order to be in radio-range at all times with all the motes\footnote{This means that the whole system described in this section was able to operate within a single WSN island}.

The motes in the wireless sensor network run the following software:
\begin{enumerate}
\item The RSSi anchors run a dedicated application for transmitting beacon messages to the robot mote, at intervals of 500 ms (time tick). This means that the mote on the robot that receives the beacons from the anchors, computes the RSSi from each anchor every time tick, and it sends to the sink all the RSSi values in a vector through a single synaptic channel.
\item The other motes run a copy of the NesC/TinyOS Communication Layer. This means that each of these motes, establish one synaptic channel with the sink to send the PIR readings.
\end{enumerate}

Finally, a basestation PC running TinyOS 2.x and the Java-based server side of the Communication and the Learning Layer was  USB-connected to the sink-mote and used to interact with the above WSN.

In Listing \ref{listing:java_hospital}, we report the pseudo-code of the base station written to program the motes in order to execute the scenario described above, in which we show how to create and start the Synaptic Channels, with their Synaptic Connections, and how to read the outputs generated by the neaural computation. 

In the first block, we configure the Synaptic Connections (that in this case correspond to the transducers to be read) for all the motes as a vector. In this case, each mote opens with the sink one Synaptic Channel with one Synaptic Connection. 

In the second block, we initialize the motes with the mote address and in the third block we create the Synaptic Channels and configure the Synaptic Connections. In particular, the mote on the robot opens one Synaptic Channel with the sink that contains a single Synaptic Connection used to transmit the RSSi values measured from all the anchors as a vector (note that with our implementation of the anchors, the RSSi can be read by the mote on the robot as any other normal transducer installed on the mote since the anchors send the beacon periodically with a dedicated software, therefore there is always a packet from which to compute the RSSi at each time tick).
Each of the PIR motes opens one Synaptic Channel with the sink, that contains one Synaptic Connection to transmit the PIR reading. 

In the fourth block, we set the time tick and we initiate the synaptic computation. Once it is started, in the fifth block, the basestation can read the output received by the sink.

The interested reader can find more information on the learning system using the synaptic connections to learn to localize the robot using RSSi in \cite{chessaRSSi14}.



%
%
%

\lstset{language=Java}
\begin{lstlisting}[caption=Pseudo-code for In-Hospital Transport Test-Bed scenario., label=listing:java_hospital]

[...]

/*--------------------------------------------*/
/* Configure transducers for all motes                       */
/*--------------------------------------------*/
Vector configuration = {RSSI, PIR, PIR, PIR, PIR, PIR, PIR};
numberOfMotes = configuration.size();
numberOfSynChannels = configuration.size();
/* in this case the number of motes corresponds to the number of Synaptic Channels because each mote opens exactly one Synaptic Channel with the Sink */

/*--------------------------------------------*/
/* Initialize motes			         			                */
/*--------------------------------------------*/
for each mote i in numberOfMotes {
	motes[i] = new Mote(MOTE_ADDRESS);
}
				
/*--------------------------------------------*/
/* Create Synaptic Channels		                            */
/*--------------------------------------------*/
for each Synaptic channel i in numberOfSynChannels {
    sc[i] = create_syn_channel_out(SINK, 1, RELIABLE);
    configure Synaptic Connections in sc[i]; //this associates the proper value of the configuration vector to the Synaptic Channel
}
/* create Synaptic Channels with only 1 Synaptic Connection each (recall that RSSI are sent all at once in a vector) with the modality realiable, that is at each time tick the readings are always sent. */
    		
/*--------------------------------------------*/
/* Set the time tick in msec	                                */
/* Start Synaptich Channels	 			                       */
/*--------------------------------------------*/
setTimeStep(500); 
start_all_syn_channels();

/*--------------------------------------------*/
/* Read received data			                   			    */
/*--------------------------------------------*/
for each Synaptic Channel sc[i] {
    for each Synaptic Connection j in sc[i] {
        data[j] = read_inputValue(sc[i], j);
    }
}

[...]

\end{lstlisting}

\subsection{AAL Test-bed}
\begin{figure}
\centering
  \includegraphics[width=14cm]{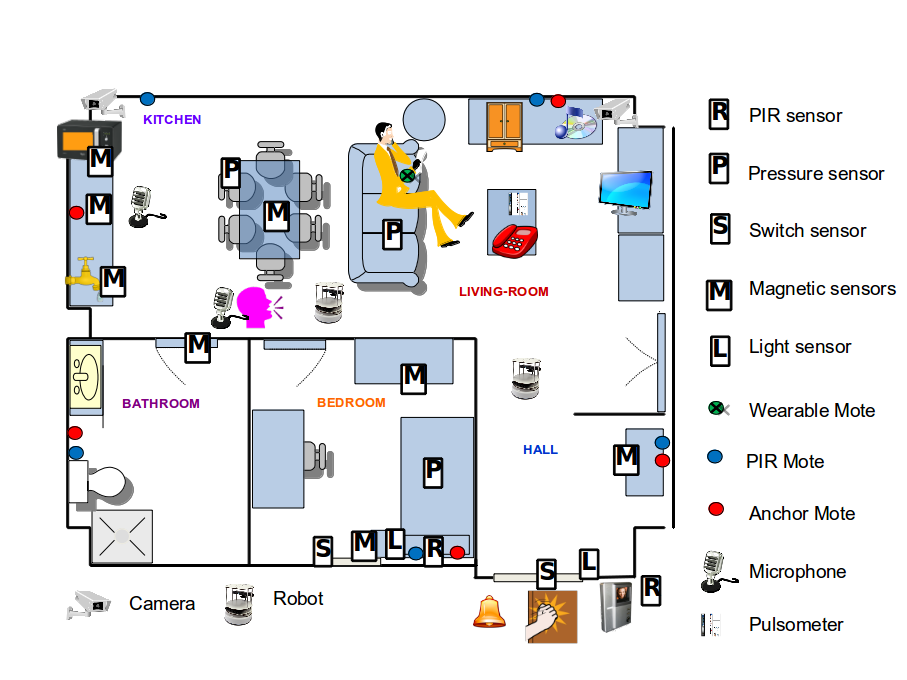}
\caption{A schematic representation of the HomeLab showing the location of
groups of sensors and actuators used in RUBICON.}\label{figure:homelab}
\end{figure}

The other test-bed used to test the RUBICON AAL system,
is the TECNALIA HomeLab, that is a fully functional 
(shown in Figure \ref{figure:homelab}) of over 45m$^2$ with a bed-room, livingroom,
kitchen area, bathroom, and a corridor. The HomeLab
is equipped with embedded sensors, effectors, and with two
mobile robots. 

The following is a list of the sensors and actuators installed in the AAL test-bed:
\begin{itemize}
\item Electronic lock and electric system for door opening at the home entrance. There is also a camera in the front door.
\item A  wireless sensor network communicating over a network based on the IEEE 802.15.4 standard. In total, we employed twelve motes based on the original open-source ``TelosB'' platform. The sensors include temperature, humidity light sensors, and PIR sensors installed in proximity of  relevant locations, such as the kitchen, and at the entrance of the other rooms in the apartment. In addition, the user wears a bracelet carrying a mote equipped with a 2-axis accelerometer sensor.
\item Konnex (KNX\footnote{KNX is approved as an International Standard (ISO/IEC 14543-3) as well as a European Standard (CENELEC EN 50090 and CENEN 13321-1) and Chinese Standard (GB/Z 20965)}) home automation system and proprietary domotic networks including light switches, blind control, alarm triggering, and general purpose input/output modules to be connected to sensors and actuators (Tv set, HiFi, etc.). Locations of sensors include drawers, cupboards and the refrigerator.
\item Network of microphones, distributed over the rooms, for sound recognition and classification: doorbell, water pouring, TV sounds, microwave bell, etc.
\end{itemize}
As in the transport scenario, each mote in this second WSN runs the nesC/TinyOS implementation of the Communication Layer and the learning modules.
For the purpose of testing the integrated and distributed system in this test-bed, we run the Learning, Control and Cognitive Layers each on a separate computer. Noticeably, at this level, intra-layer interaction occurs over the LAN and thus is not subjected to the bandwidth or computational constraints imposed by battery-operated WSN motes. Finally, an additional computer runs an instance of a proprietary home automation middleware, which is used to collect sensor readings from the KNX sensors and from the network of microphones.

An important aspect of this scenario is that, we developed a Proxy aimed at providing to the Java implementation of the \prog Network level a transparent communication mechanism working with heterogeneous networks. In this specific test-bed, the Proxy enables the Control Layer to monitor and control sensors and actuators in both the WSN and the KNX network. In \prog the Proxy acts as access point over the PEIS tuplespace for all the sensors and actuators in the system - no matter which hardware and communication protocol are in use. Specifically, the Proxy makes available the data gathered by the sensors in the underlying networks, to the higher layers (e.g. control and learning components), which do not need to account for specific network protocols. In addition, they can simply post configuration instructions, for both sensors and actuators, and see those instructions automatically translated by the Proxy in the specific network protocol and routed to their intended device.

\begin{figure}
\centering
  \includegraphics[width=14cm]{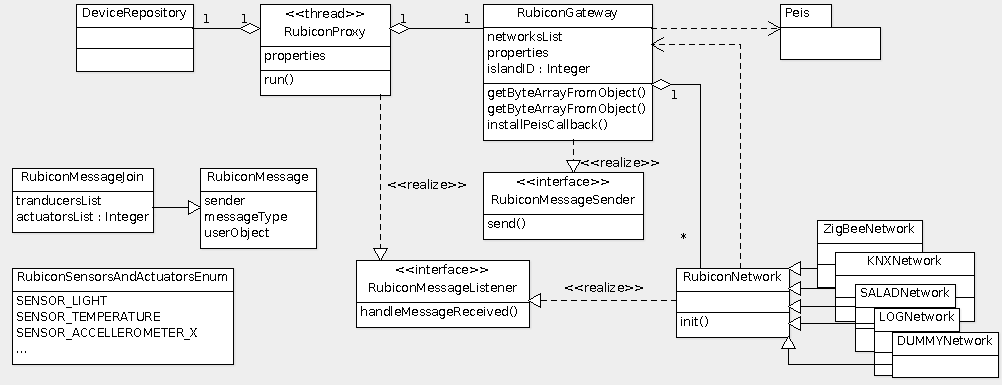}
\caption{Proxy system: Class Diagram.}\label{figure:umlproxy}
\end{figure}

Figure \ref{figure:umlproxy} shows  a UML class diagram representing the main classes and the interfaces included in the Java-based implementation of the Proxy. Each new network must subclass the class RubiconNetwork (the Java class that implements the Network level) in order to access configuration utilities and to communicate with the Proxy. Messages between the network implementation level and the higher levels of the architecture,
and between multiple and distributed higher-level components are encapsulated by object instances of the RubiconMessage class, representing the source of each message and its content. Higher level components can define their own specialized version of the RubiconMessage (e.g. to carry application or layer-specific information),
For instance, the Proxy uses a TCP-IP adapter in order to interact with a middleware used for integrating KNX devices and to publish in PEIS the sensor data collected from the KNX network.
The proxy handles PEIS subscriptions to data gathered from the WSN, which are fulfilled by (i) activating and configuring (through messages routed by the sink mote) the sampling of sensor data, (ii) interpreting the resulting data updates once they are received by the WSN sink node, and (iii) converting them into a standard format and posting them as tuples in the PEIS tuplespace.

In order to support dynamic networks, with devices joining and leaving the system at run-time, for instance, due to
device mobility, component failure, and power outage, a simple discovery protocol is defined to allow the Network
level to signal the presence of new devices, together with the description of the type of the sensors and actuators
installed on each device. In the case of WSN, the protocol must be initiated at the device level. For
WSN networks, RUBICON relies on the embedded nesC/TinyOS implementation of the Communication
Layer described in Section \ref{sec:design}.

\lstset{language=Java}
\begin{lstlisting}[caption=Java code for the proxy for translating TinyOS messages through PEIS., label=listing:java_proxy]
@Override
public void init() throws Exception {
	// Install classes to convert between TinyOS messages and RubiconMessage objects
	installMessageConverter(new RubiconMessageJoinConverter());
	installMessageConverter(new RubiconMessageCtrlActuateCmdConverter());
	installMessageConverter(new RubiconMessageCtrlPeriodicCmdConverter());
	installMessageConverter(new RubiconMessageUpdatesConverter());
	
	// Initialise Network			
	Network.getNetwork()._init(this.getIslandID(), this.getGateway(), this.getProperties());
	
	// Gets Connectionless API	
	connection  = Connectionless.getConnectionless();

	// Subscribes to Control messages
	connection.setOnReceiveConlessControlMessage(
		new ReceiveConlessControlListener() {

			@Override
			public void onReceiveConlessControlMessage(RubiconAddress sender, byte[] payload, short nbytes) {
				Message msg = null;
				switch(payload[0]) {
					case Definitions.AM_SERIAL_JOINED_MSG:
						msg = new SerialJoinedMsg(payload);
						break;
					case Definitions.AM_UPDATES_MSG:
						msg = new UpdatesMsg(payload);
						break;
					}
					
					if (msg != null) {
						// translate message
						translateAndSendToGateway(msg, sender, new RubiconAddress(getIslandID(),-1));
					}
					
				}						
			}
	);

}
\end{lstlisting}


With such a setup, the Proxy can be used to: (i) read data gathered from sensors installed in the environment, such as switch sensors signalling when the drawers/doors are open or closed, and occupancy sensors signalling when the user moves in certain areas, (ii) send instructions to effectors, such as lights, blinds, door locks and appliances, (iii) sense the status of these effectors and know when the user interacts with them (i.e. when he/she manually switches on/off the TV, lights, etc), (iv) recognize when new sensors are added or existing ones are removed, and notify these events to all the higher layers of \prog architecture.

In order to show the easiness of use of our software, in Listing \ref{listing:java_proxy}, we report the Java code for configuring the proxy to transparently translate TinyOS messages into PEIS messages. The first block loads the classes used to convert the TinyOS messages. The second block initializes the network object with the island ID and the instance of the gateway. The third block gets the instance to the connectionless object to access the API for sending and receiving messages through the Connectionless Transport-Layer component. Finally, the last block registers the proxy component to the connectionless messages coming from the sensor network and implements the translation of incoming messages. In particular, upon receiving a TinyOS message, the proxy converts the message to the proper Java class according to the first byte of the message, and then forwards the translated message to the Gateway.

In order to feed the neural network with data streams useful for the learning phase,
we used a data-logger tool developed in \prog project \cite{amato2013data} for remotely control a set of motes for both real-time and off-line data acquisition.
A data benchmark for the assessment of human activity recognition in the presented scenario is provided in \cite{amato2016benchmark}.